\documentclass[conference]{IEEEtran}
\IEEEoverridecommandlockouts
\usepackage{cite}
\usepackage{amsmath,amssymb,amsfonts}
\usepackage{graphicx}
\usepackage{textcomp}
\usepackage{xcolor}
\def\BibTeX{{\rm B\kern-.05em{\sc i\kern-.025em b}\kern-.08em
    T\kern-.1667em\lower.7ex\hbox{E}\kern-.125emX}}
    
\usepackage{hyperref}
\usepackage{tikz}
\usepackage{amsmath}
\usepackage{graphicx}
\usepackage{multirow}
\usepackage{subfigure}
\usepackage{filecontents}
\usepackage{footnote}
\usepackage{listings}
\usepackage{xspace}
\usepackage{booktabs}
\usepackage[ruled,vlined,linesnumbered]{algorithm2e}
\SetKwComment{Comment}{$\triangleright$\ }{}
\SetKwInOut{Variables}{Variables}
\usepackage{algorithm2e}
\usepackage{soul}
\usepackage{algorithmicx}  
\usepackage{algpseudocode}  
\usepackage{amssymb}
\usepackage{pifont}
\usepackage[flushleft]{threeparttable}
\usepackage{xcolor}
\usepackage{makecell}
\usepackage{lipsum}
\usepackage{balance}
\usepackage{caption} 
\usepackage{enumitem}
\setlist{noitemsep}
\usepackage{url}

\definecolor{firebrick}{rgb}{0.7, 0.13, 0.13}
\definecolor{darkblue}{rgb}{0,0,0.55}
\definecolor{grey}{rgb}{0.8,0.8,0.8}

\newcommand\nrule[1][black]{\textcolor{#1}{\rule{3mm}{2mm}}}
\newcommand{\firebrick}[0]{\nrule[firebrick]}
\newcommand{\darkblue}[0]{\nrule[darkblue]}

\newcommand{\one}{({\em i}\/)}
\newcommand{\two}{({\em ii}\/)}
\newcommand{\three}{({\em iii}\/)}
\newcommand{\four}{({\em iv}\/)}



\usepackage{siunitx}
\newcommand{\pie}[1]{
\begin{tikzpicture}
 \draw (0ex,0ex) circle (1ex);
 \fill (0ex,-1ex) arc (-90:(#1-90):1ex) -- (0ex,-1ex) -- cycle;
\end{tikzpicture}
}

\newcommand{\rect}[1]{
\begin{tikzpicture}
 \draw (-1ex,-1ex) rectangle (1ex,1ex);
 \fill (#1ex,-1ex) rectangle (1ex,1ex);
\end{tikzpicture}
}

\newcommand{\rectgrey}[1]{
\begin{tikzpicture}
 \draw (-1ex,-1ex) rectangle (1ex,1ex);
 \fill [color=grey] (#1ex,-1ex) rectangle (1ex,1ex);
\end{tikzpicture}
}

\def\eg{\emph{e.g.,}\xspace}

\def\ie{\emph{i.e.,}\xspace}
\def\etal{\emph{et al.}\xspace}

\DeclareMathOperator{\init}{\texttt{init\_style\_set}}
\DeclareMathOperator{\len}{length}
\DeclareMathOperator{\colorprox}{\texttt{color\_prox}}
\DeclareMathOperator{\none}{None}
\DeclareMathOperator{\score}{\texttt{target\_score}}
\DeclareMathOperator{\transfer}{\texttt{style\_transfer}}
\DeclareMathOperator{\NES}{\texttt{NES}}
\DeclareMathOperator{\PGD}{\texttt{PGD}}
\DeclareMathOperator{\BPGD}{\texttt{BPGD}}
\DeclareMathOperator{\tovideo}{\texttt{to\_video}}
\DeclareMathOperator{\argmin}{arg\,min}

\DeclareMathOperator{\R}{\mathbb{R}}

\newcommand{\nParticipants}[0]{\num{160}\xspace}

\begin{document}

\title{\Large \bf StyleFool: Fooling Video Classification Systems via Style Transfer}

\author{
\IEEEauthorblockN{
Yuxin Cao\IEEEauthorrefmark{1}, 
Xi Xiao\IEEEauthorrefmark{1}, 
Ruoxi Sun\IEEEauthorrefmark{2}, 
Derui Wang\IEEEauthorrefmark{2},
Minhui Xue\IEEEauthorrefmark{2} and 
Sheng Wen\IEEEauthorrefmark{3}}
\IEEEauthorblockA{
\IEEEauthorrefmark{1}Shenzhen International Graduate School, Tsinghua University, China}
\IEEEauthorblockA{
\IEEEauthorrefmark{2}CSIRO's Data61, Australia}
\IEEEauthorblockA{
\IEEEauthorrefmark{3}Swinburne University of Technology, Australia}
}

\maketitle

\begin{abstract}
Video classification systems are vulnerable to adversarial attacks, which can create severe security problems in video verification. Current black-box attacks need a large number of queries to succeed, resulting in high computational overhead in the process of attack. On the other hand, attacks with restricted perturbations are ineffective against defenses such as denoising or adversarial training. In this paper, we focus on unrestricted perturbations and propose \textit{StyleFool}, a black-box video adversarial attack via style transfer to fool the video classification system. StyleFool first utilizes color theme proximity to select the best style image, which helps avoid unnatural details in the stylized videos. Meanwhile, the target class confidence is additionally considered in targeted attacks to influence the output distribution of the classifier by moving the stylized video closer to or even across the decision boundary. A gradient-free method is then employed to further optimize the adversarial perturbations. We carry out extensive experiments to evaluate StyleFool on two standard datasets, UCF-101 and HMDB-51. The experimental results demonstrate that StyleFool outperforms the state-of-the-art adversarial attacks in terms of both the number of queries and the robustness against existing defenses. Moreover, 50\% of the stylized videos in untargeted attacks do not need any query since they can already fool the video classification model. Furthermore, we evaluate the indistinguishability through a user study to show that the adversarial samples of StyleFool look imperceptible to human eyes, despite unrestricted perturbations. 

\end{abstract}


\section{Introduction}
With short videos becoming more and more popular in today's era, videos have gradually become inevitable in people's daily lives~\cite{papernot2016transferability}. To date, a large number of video-sharing mobile applications have sprung up, such as Tiktok, Facebook Watch, and Youtube~\cite{kong2018will,cheng2013understanding,video_cisco}. In order to avoid politically sensitive disputes and protect the physical and mental health of minors, it is necessary to verify the videos uploaded by users to prevent the spread of illegal or criminal videos such as violence, pornography, and malicious marketing~\cite{liu2020video, marciel2016understanding}. However, the rapid increase in the number of videos and the limitations of human and time resources have become challenges to manual verification. Therefore, the demand for machine-learning-assisted video classification systems increases greatly~\cite{liu2020video}. 

Severe consequences may occur once the video verification is compromised: an attacker can maliciously modify pixels of an illegal video, \eg a violent or pornographic video, to bypass the video verification classifier (\ie the video is falsely classified as benign) and expose the video to the public. 
If the illegal video is further circulated widely, it may cause public panic and other adverse effects, such as the threats to video content rating for children and the risk of video spreading for terrorist purposes in social networks~\cite{yuan2019stealthy,child_porn}. 
In addition, the emerging technologies in artificial intelligence, such as DeepFakes~\cite{deepfakes} and Face2Face~\cite{thies2016face2face}, extremely reduce the difficulty to generate fake videos, making video verification more important than ever. 
This is substantiated by the recent fact that a fake video of the Ukrainian president calling on his soldiers to lay down their weapons was uploaded to a Ukrainian news website~\cite{wakefield2022deepfake}.
There is no doubt that the robustness of video classification systems is security-critical. It is essential for the security research community to investigate potential attacks against robustness and thoroughly evaluate the machine-learning-assisted system before full deployment. 

\begin{table*}[t]  
\centering
\caption{A comparison of adversarial attacks against video classifiers.}
\label{tab:study_comparison}
\resizebox{0.99\linewidth}{!}{
\begin{threeparttable}
\begin{tabular}{c|cc|cc|cc|cc|cc|cccc}
\toprule
\multirow{2}{*}{\textbf{Approaches}} 
& \multirow{2}{*}{\textbf{Online}} 
& \multirow{2}{*}{\textbf{Offline}}
& \multirow{2}{*}{\textbf{White-box}} 
& \multirow{2}{*}{\textbf{Black-box}}
& \multirow{2}{*}{\textbf{Universal}} 
& \multirow{2}{*}{\textbf{One-on-one}} 
& \multirow{2}{*}{\textbf{Untargeted}} 
& \multirow{2}{*}{\textbf{Targeted}}
& \multirow{2}{*}{\textbf{Restricted}}
& \multirow{2}{*}{\textbf{Unrestricted}} 
& \multicolumn{4}{c}{\textbf{Compromised Defenses}}\\
\cline{12-15}
& & & & & & & & & & & AdvIT~\cite{xiao2019advit} & Comdefend~\cite{jia2019comdefend,jia2019identifying} & RS~\cite{jeremy2019certified} & AT~\cite{goodfellow2014explaining,ali2019adversarial} \\
\midrule
C-DUP~\cite{li2019stealthy} & \pie{360} & \pie{0} & \pie{360} & \pie{0} & \pie{360} & \pie{0} & \pie{360} & \pie{0} & \pie{360} & \pie{0} & \rectgrey{-1} & \rectgrey{-1} & \rectgrey{-1} & \rectgrey{-1} \\
U3D~\cite{xie2022universal} & \pie{360} & \pie{0} & \pie{0} & \pie{360} & \pie{360} & \pie{0} & \pie{360} & \pie{0} & \pie{360} & \pie{0} & \rect{-1} & \rectgrey{-1} & \rect{-1} & \rect{-1} \\
\midrule
V-BAD~\cite{jiang2019black} & \pie{0} & \pie{360} & \pie{0} & \pie{360} & \pie{0} & \pie{360} & \pie{360} & \pie{360} & \pie{360} & \pie{0} & \rect{1} & \rect{1} & \rectgrey{1} & \rectgrey{-1} \\
H-Opt~\cite{wei2020heuristic} & \pie{0} & \pie{360} & \pie{0} & \pie{360} & \pie{0} & \pie{360} & \pie{360} & \pie{360} & \pie{360} & \pie{0} & \rect{1} & \rect{1} & \rectgrey{1} & \rectgrey{-1} \\
\makecell[c]{\textbf{StyleFool (ours)}} & \pie{0}\tnote{1} & \pie{360} & \pie{0} & \pie{360} & \pie{0}\tnote{1} & \pie{360} & \pie{360} & \pie{360} & \pie{0} & \pie{360} & \rect{-1} & \rect{-1} & \rect{-1} & \rect{-1} \\
\bottomrule
\end{tabular}
\begin{tablenotes}
\item[]\pie{360}: the item is supported by the attack; \pie{0}: the item is not supported by the attack. 
\item[]\rect{-1}: the attack can compromise the defense; \rect{1}: the attack cannot compromise the defense; \rectgrey{-1}: not mentioned in the corresponding research.
\item[1] Although StyleFool proposed in this paper focuses on offline one-on-one attacks, we argue that our framework can also produce universal perturbations which is suitable for online attacks. See more details in the discussion.
\end{tablenotes}
\end{threeparttable}
}
\vspace{-1mm}
\end{table*}

As a family of machine learning, Deep Neural Networks (DNNs) have become an indispensable tool in the field of multimedia~\cite{krizhevsky2012imagenet,ji20123d,carreira2017quo,carlini2016hidden,chen2018deep}. 
Despite significant advantages and far-reaching influence, studies have found that DNNs could be vulnerable to adversarial attacks~\cite{szegedy2013intriguing,goodfellow2014explaining}. 
More specifically, by superimposing an elaborately designed perturbation (even unsuspicious or imperceptible to human eyes) on an input sample, an attacker can fool the classifier into misclassification~\cite{goodfellow2014explaining,carlini2017towards}. 
Recent research found that, as long as a pixel of the input image is perturbed, the classifier can be fooled successfully~\cite{su2019one}. 
The resulting security problem has encountered enormous challenges with the application of DNNs in many aforementioned fields. 

In contrast to a plethora of studies focusing on image adversarial attacks~\cite{goodfellow2014explaining, carlini2017towards, moosavi2016deepfool, papernot2016limitations, papernot2017practical, bhagoji2018practical, chen2017zoo, ilyas2018black, cheng2018query, zajac2019adversarial, yuan2019stealthy}, research in the video domain has been slowly ramping up~\cite{wei2019sparse,li2019stealthy,jiang2019black,wei2020heuristic,xie2022universal,li2021geometric,wang2021reinforcement,chang2022adversarial}. One of the main reasons is that videos contain temporal information, which greatly increases the attack difficulty. Table~\ref{tab:study_comparison} shows the comparison of some existing video attacks, which can be categorized into two branches: \emph{universal attacks} and \emph{one-on-one attacks}. 
The attacks can be launched in either \emph{online} scenario or \emph{offline} scenario.
The online scenario requires real-time attacks adding perturbations to online videos with minimal latency.
In contrast, the offline scenario has a minimal concern towards the attacking overhead since the attacker has unlimited local access to the victim videos and models.
Universal attacks such as C-DUP~\cite{li2019stealthy} and U3D~\cite{xie2022universal} aim to generate a universal perturbation for all videos so as to fool the classifier. 
One-on-one attacks such as V-BAD~\cite{jiang2019black} and H-Opt~\cite{wei2020heuristic} aim to produce a sample-specific perturbation for each input video. 
Universal perturbations are only available in untargeted attacks, while one-on-one perturbations can be applied to both targeted and untargeted attacks. 
From the perspective of the adversary's knowledge, white-box attacks (\eg C-DUP~\cite{li2019stealthy}) have a stronger assumption that the adversary can access the training data and the classifier model. This makes white-box attacks not as practical as black-box attacks.
Additionally, state-of-the-art video attacks, which introduce imperceptible $\ell_p$-norm restricted perturbations, are shown to be successfully defended by adversarial defenses~\cite{xie2019feature}, such as AdvIT~\cite{xiao2019advit}, Comdefend~\cite{jia2019comdefend,jia2019identifying}, and Randomized Smoothing (RS)~\cite{jeremy2019certified}. 
GAN-generated images can also be distinguished by CNN-generated image detection~\cite{wang2020cnn-generated}. 
Furthermore, we point out that the existing one-on-one attacks~\cite{jiang2019black,wei2020heuristic} consume a large number of queries (approximately $(5 \sim 26) \times {10^4}$) when generating adversarial perturbations. 

In summary, we list three limitations of the state-of-the-art video attacks as follows.
\one~White-box attacks are less practical than black-box attacks for heavily overestimating the capability of the adversary.
\two~A large number of queries are needed in the existing video attacks~\cite{jiang2019black,wei2020heuristic}, resulting in high complexity especially when the target model is a commercial system.
\three~Restricted perturbations can be defended by adversarial defenses successfully. 

In this research, 
we propose \textit{\textbf{StyleFool}}, a black-box unrestricted adversarial attack framework against video classification systems, that has the following advantages. \one~StyleFool introduces \textit{unrestricted} perturbations which transfer videos into another style while preserving the semantic information in the black-box setting. Concretely, we focus on changing the non-semantic critical information which will not confuse human understanding of video content, but mislead the classifier. We argue such perturbations are observable but imperceptible that preserve the indistinguishability of original videos and break the limitation existing in the state-of-the-art attack frameworks. 
\two~Style transfer can initialize powerful perturbations. Pre-processing videos by adversarial style transfer can significantly reduce the number of queries in the subsequent attack. 
\three~Unrestricted perturbations with high temporal consistency are considered, which enhances the robustness against state-of-the-art defenses, including detection defense (\eg AdvIT~\cite{xiao2019advit}), reconstruction defense (\eg ComDefend~\cite{jia2019comdefend}), certified defense (\eg RS~\cite{jeremy2019certified}), and a CNN-generated detector~\cite{wang2020cnn-generated}. The source code is publicly available at \url{https://github.com/JosephCao0327/StyleFool}.

\begin{figure*}[t]
  \centering
  \includegraphics[width=0.85\linewidth]{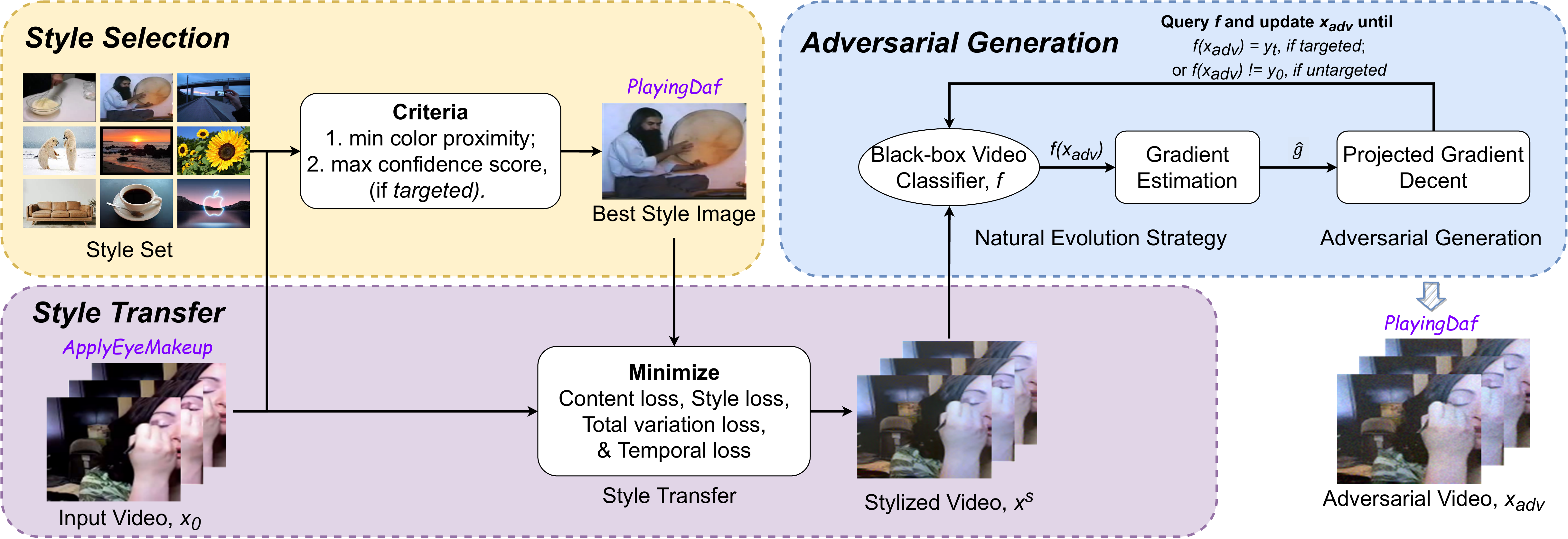}
  \caption{An overview of StyleFool against video classification systems.}
  \label{fig:overview}
\end{figure*}

\noindent \textbf{Our main contributions are as follows.}
\begin{itemize}[leftmargin=*]
\item We propose a novel attack framework, StyleFool, in video domain. To the best of our knowledge, StyleFool is the \textit{first} attempt to attack video classification systems with style-transfer-based unrestricted perturbations.

\item We initialize perturbations with adversarial style transfer, which pushes the stylized video towards the decision boundary to reduce the queries required during adversarial sample generation. Meanwhile, the temporal consistency and indistinguishability (\ie whether an adversarial video can be distinguished from the original video by human subjects) of the video are preserved.

\item 
Our extensive experiments indicate that StyleFool 
effectively moves the videos to the vicinity of decision boundaries.
It also reduces the number of queries, compared to the state-of-the-art video attacks, V-BAD~\cite{jiang2019black} and H-Opt~\cite{wei2020heuristic}, by 43\% and 83\%, respectively. A user study demonstrates the indistinguishability of StyleFool. We show that StyleFool is an efficient video attack framework, which is conducive to improving the robustness of the video classification model.

\item 
We demonstrate the capability of StyleFool to bypass the state-of-the-art defense strategies, AdvIT~\cite{xiao2019advit}, ComDefend~\cite{jia2019identifying}, RS~\cite{jeremy2019certified}, {and CNN-generated image detection~\cite{wang2020cnn-generated}}. We also discuss and recommend the potential mitigation. 
We believe that our research will arouse community's attention of such style-transfer-based attacks in security scenarios, especially in video verification.

\end{itemize}

\noindent \textbf{Ethical considerations.~}
The Human Research Ethics Committee of the authors' affiliation determined that the study was exempt from further human subjects review, and we followed the best practice for ethical human subjects survey research, \eg all questions were optional and we did not collect unnecessary personal information. All participants are over 18 years old and they consented for their answers to be used for academic research.

\section{Background and Threat Model}\label{sec:background}
In this section, we first introduce the background information and existing research related to our work. Then we describe the threat model from three aspects: adversary’s goals, capabilities, and knowledge.

\subsection{Background and Related Work}
\noindent \textbf{DNN-based video classification systems.~} 
A video classification system automatically classifies videos into corresponding labels based on the semantic content of human behaviors in complex events with DNNs, such as convolutional neural networks (CNNs). Several DNN-based video classification systems, such as TSN~\cite{simonyan2014two-stream}, C3D~\cite{tran2015learning}, LRCN~\cite{donahue2015long-term}, CNN+LSTM~\cite{ng2015beyond}, and I3D~\cite{carreira2017quo}, have been proposed for video recognition tasks. Among them, by incorporating temporal information into a two-dimensional CNN (2D-CNN)~\cite{karpathy2014large, simonyan2014two-stream}, a 3D-CNN network was proposed in C3D to extract spatiotemporal features of videos. Inspired by 3D ConvNets~\cite{ji20123d}, TSN~\cite{simonyan2014two-stream} and CNN+LSTM~\cite{ng2015beyond}, I3D was proposed with two-stream inflated filters for video classification. 
C3D and I3D perform better than other models in video classification tasks and are prevalently adopted in various applications~\cite{tran2015learning, carreira2017quo}.
On the flip side, the security issues in C3D and I3D are spotlighted due to their popularity~\cite{li2019stealthy,jiang2019black,wei2020heuristic,xie2022universal}.

\noindent \textbf{Adversarial attacks and defenses.~} 
Existing studies in the field of adversarial attack mainly focus on the image domain. Most of them are based on $\ell_p$-norm perturbations~\cite{goodfellow2014explaining, carlini2017towards, moosavi2016deepfool, papernot2016limitations, papernot2017practical, bhagoji2018practical, chen2017zoo, ilyas2018black, cheng2018query, zajac2019adversarial, yuan2019stealthy}, while some recent studies have considered unrestricted attacks, breaking the boundaries in image attacks~\cite{hosseini2018semantic,bhattad2020unrestricted,shamsabadi2020colorfool,yang2020patchattack,duan2020camouflage,brown2018unrestricted,song2018constructing}. For example, ColorFool~\cite{shamsabadi2020colorfool} changed the color of the image and Patchattack~\cite{yang2020patchattack} added a patch to the image. 
It has been verified that the attack performance of unrestricted attacks is better than that of restricted attacks~\cite{shamsabadi2020colorfool}, so as the circumvention performance against defenses~\cite{yang2020patchattack}. 

Adversarial attacks against video classification can be classified as either white-box attacks~\cite{li2019stealthy,wei2019sparse,inkawhich2018adversarial,pony2021flickering,chen2021appending,chang2022adversarial} or black-box attacks~\cite{jiang2019black,wei2020heuristic,xie2022universal,wang2021reinforcement,li2021geometric,Kumar2020finding}. The first attack towards video classifiers was operated in a white-box setting by optimizing perturbations bounded by $\ell_1$ or $\ell_2$ norm~\cite{wei2019sparse}. Inspired by Generative Adversarial Networks (GANs), an offline universal adversarial perturbation called C-DUP was proposed for white-box attacks~\cite{li2019stealthy}. One recent work was devoted to attacking the video compression model in the context of white-box setting, and the method was further exploited on the classification model for the adversarial attack~\cite{chang2022adversarial}. V-BAD was proposed to launch attacks with limited queries against black-box classifiers~\cite{jiang2019black}. Subsequently, in light of the Opt attack~\cite{cheng2018query}, a heuristic attack was proposed to increase the perturbation sparsity in black-box attacks by only adding perturbations to salient regions in key frames~\cite{wei2020heuristic}. To enhance the transferability of black-box attacks, a universal three-dimensional perturbation (U3D) was lately proposed~\cite{xie2022universal}. 

The state-of-the-art defenses detect or remove adversarial perturbations in the videos~\cite{xiao2019advit, jia2019identifying}. AdvIT detects inconsistencies between video frames and frame-wise optical flows to expose adversarial videos~\cite{xiao2019advit}. Furthermore, Comdefend relies on compressing and reconstructing a video with latent noise to mitigate the effect of adversarial perturbations ~\cite{jia2019identifying}. In the image domain, RS is proved to be effective for $\ell_2$-norm attacks~\cite{jeremy2019certified}.

\noindent \textbf{Video style transfer.~} 
Style transfer methods stylize a content image according to a style image, such as an artwork by a famous painter.
Hertzmann~\etal~\cite{hertzmann2001image} first explored image style transfer using image analog. Subsequently, it was found that content loss and style loss can better preserve the similarity between an input image and the stylized image in the feature space~\cite{a.2015a}. Recently, the total variance loss was introduced to further ensure the smoothness of images at the pixel-wise level~\cite{justin2016perceptual}. 
However, when applying style transfer to videos, the inconsistency between adjacent frames becomes a troublesome issue~\cite{huang2017real,gao2018reconet}. To ease the problem, the temporal loss was further developed to generate consistent and stable stylized video sequences~\cite{ruder2016artistic}. 

\noindent \textbf{Adversarial attack using style transfer.~}
AdvCam~\cite{duan2020camouflage} first applied style transfer to image adversarial attacks by applying style transfer on specific areas in an image. However, in video scenarios, the most challenging point is to generate an adversarial stylized video while keeping the consistency of each frame in the video. What attacking video classifiers by style transfer differs from AdvCam is that, it is challenging to combine the loss of style transfer and adversarial attack, since the learning rates and the times of iteration in the two steps are rather different when learning video samples. 
Further, we cannot only select some specific areas in frames for video style transfer, considering that the style of an area is likely to change in a video (\eg the motion of an actor).
We argue that it is difficult to squarely extend state-of-the-art style transfer attacks from images to videos.

\subsection{Threat Model}
We detail the threat model of StyleFool against video classification systems in this section.

\noindent \textbf{Adversary's goals.} 
Given a DNN $f\left( {x} \right):x \to y$, it takes a video $x\in {\mathbb{R}^{T \times H \times W \times C}}$ from a video set $X$ as input. Herein, $T$, $H$, $W$, and $C$ are the frame number, height, width, and channel number of $x$, respectively.
The model outputs a $K$-class prediction label $y \in \{1,...K\},\ y\in Y$, where $Y$ is the set of predicted labels of $X$. 
The adversary's goal is to find an adversarial counterpart ${x_{adv}}$ of $x$ satisfying the constraints listed in Equation~\ref{equ:goal}. The output label is a predetermined target class $y_t$ in targeted attacks.
Otherwise, in untargeted attacks, $x_{adv}$ is misclassified into a random label other than the ground truth label $y_0$ of an input video~$x$.
\begin{equation}\label{equ:goal}
\left\{ \begin{array}{l}
f(x_{adv}) = y_t, ~~ if~targeted,\\
f({x_{adv}})\neq y_0, ~~ if~untargeted.
\end{array} \right.
\end{equation}

\noindent \textbf{Adversary's capabilities.}
In our research, we limit the adversary's capabilities as follows. The adversary is capable of crafting perturbations and superimposing them on given videos in the offline setting.
Please note that we only focus on one-on-one attacks which crafts sample-specific perturbations. Compared with universal attacks, one-on-one attacks can achieve both untargeted and targeted attacks.

To establish a video frame set from which the style images could be selected, the adversary may collect a large number of videos from publicly available datasets or other online resources. 
Furthermore, in the targeted attack scenario, when the style images from the targeted class are not satisfying, the adversary can search and download images or videos according to common sense about the target class to form the style set.

\noindent \textbf{Adversary's knowledge.~}
Different from the black-box setting in attacks towards images~\cite{chen2017zoo,brendel2017decision,cheng2018query} and other state-of-the-art work in videos~\cite{wei2020heuristic}, we place stricter restrictions on the knowledge possessed by the adversary, \ie the adversary can neither access the attacked model parameters nor the training set. The adversary can only access the top-1 label $y$ (the label with the highest confidence score) and its confidence score $p(y|x)$. 
An upper limit bound of attacker's budget on the number of queries is set; otherwise, in offline scenarios, the attack will always succeed if the query resources are not limited. Such a query limit also simulates the real-world black-box attacks, especially when the attacking targets are commercial video classifiers.
Specifically, we adopt the \emph{query-limited partial information} setting~\cite{ilyas2018black} in our research.

\section{Design of StyleFool}\label{sec:methodology}
The framework of StyleFool is depicted in Figure~\ref{fig:overview}. 
StyleFool can be divided into three stages which are style selection, style transfer, and adversarial sample generation.
At first, according to the style selection criteria (\ie the color theme proximity and the target class confidence), the best style is selected for style transfer. 
Then, by considering content, style, total variance, and temporal loss terms, the clean video is transferred into the selected style. 
Finally, a black-box adversarial attack is conducted to consolidate perturbations that can fool the target model. 

\subsection{Style Selection}\label{sec:style selection}
\begin{figure}[t]
    \centering
    \includegraphics[width=0.9\linewidth]{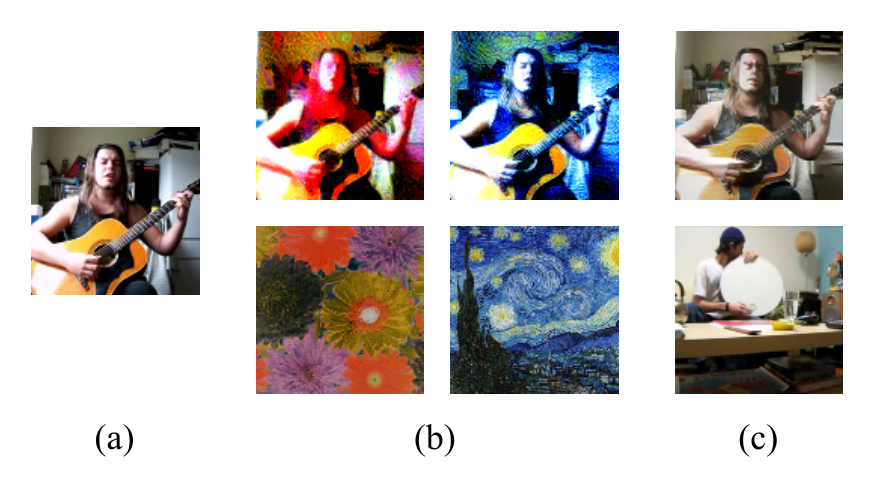}
    \caption{Examples of stylized images: (a) the original image; (b) style transfer results using artistic styles; (c) style transfer result of our approach and the selected style. The introduction of color theme proximity produces more indistinguishable style transfer, compared with the use of artistic styles.}
    \label{fig:abstract_styles}
\end{figure}

Style selection serves for two purposes. 
First, it searches for style images that ensure the stylized videos are indistinguishable from the original videos. Moreover, properly selected style images help initialize the stylized video in the vicinity of the decision boundary to reduce the number of queries in the subsequent attacking process.
As shown in Figure~\ref{fig:abstract_styles}, the existing style transfer approaches~\cite{huang2017real, ruder2016artistic,gao2018reconet} aim to conduct style transfer based on artistic styles that are distinct from the styles of the original images.
Without a careful selection of style images, style transfer may produce changes that can be easily captured by human eyes. 
This is in contrast with the aim of adversarial attacks regarding the indistinguishability of perturbations.
To preserve the indistinguishability of the stylized videos, we propose a collective set of style selection approaches for both targeted and untargeted attacks.

\subsubsection{Style Selection Based on Color Proximity}
We first quantize pixel values before calculating the color proximity.

\noindent \textbf{Median Cut (MC).~} 
MC resembles a naive recursive clustering method that computes centroids of pixel values which will next serve as color themes for calculating color proximity~\cite{heckbert1982color}. 
In practice, the first frame of a clean video is selected to produce the color themes of the video. 
Since the target model classifies short videos filming single human actions, the first frames of the video could be visually similar to other frames. 
This selection scheme is proved effective through our experiments. 
The medians are used as the boundary to ensure that the pixels are divided into two subsets of the same size and can be separated further. 
Concretely, given an image $z$ of $n$ pixels, we let it be a set $z:=\{z^R, z^G, z^B\}$ of pixel sets grouped by the RGB channels.
Note that $z^c\subset\mathbb{R}^+,\ c\in\{R, G, B\}$, where $z^c$ is the set of pixels in channel $c$. 
$\underline{z^c}=\inf\ {z^c}$ and $\overline{z^c}=\sup\ {z^c}$ are the infimum and the supremum of $z^c$, respectively. 
To obtain $m$ color themes, first, 1) the pixels in $z$ are sorted in ascending order of their pixel values in a channel $c$ satisfying $c=\arg\max|\overline{z^c}-\underline{z^c}|$. 
Next, 2) the pixels are separated into two half sets $z_1 = \{z_{1,1},...z_{1,{n/2}}\},\ z_1\leq median(z^c)$ and $z_2=\{z_{2,n/2+1},...z_{2,n}\},\ z_2 > median(z^c)$ based on the median value $median(z^c)$. 
MC repeats 1) and 2) on $z_1$ and $z_2$ and their half sets till $m$ sets of pixels are generated. 
Finally, the medians of the $m$ pixel sets are returned as $m$ quantized color themes of $z$.

\noindent \textbf{Color Proximity.~}
We define the color proximity between a video and its style image as their Euclidean distance of color themes. 
A smaller proximity indicates a better style image with which style transfer can induce less perceptible changes in the original videos.
In lieu of calculating the proximity in the RGB space, we turn to the HSV representation which aligns better with human perception~\cite{smith1978color}.
We first project the RGB values obtained from MC into the HSV space. Next, the color proximity is measured in the HSV-projected XYZ space (see detailed HSV-to-XYZ transformation in Appendix~\ref{appendix_hsv}). 
Given the XYZ coordinate $\phi^x_{i}$ of the $i$-th color theme in the clean video $x$ and $\phi^s_{j}$ of the $j$-th color theme in the style image $s$, the \emph{one-versus-one} color proximity of $s$ with respect to $x$ is expressed as
\begin{equation}\label{eq:cp}
d^{s}_{i,j} = {\left\| {\phi^x_{i} - \phi^s_{j}} \right\|_2}.
\end{equation}

Furthermore, it should be noticed that the number of color themes affects the indistinguishability of the output videos.
An insufficient number of color themes may lead to generating videos with
unnatural content. 
This concern is emphasized in videos having a dominant but semantically less important background since the color themes are dominated by the background in most images. Therefore, as shown in Equation~\ref{equ:color_sum}, we consider the \emph{$N_c$-versus-$N_c$} color proximity between $x$ and $s$. 
The style image $s^*$ with the minimal \emph{$N_c$-versus-$N_c$} color proximity is selected as the style image in style transfer. 
\begin{equation}\label{equ:color_sum}
{s^*} = \mathop {\arg \min }\limits_{s \in {S_t}} \left( {\sum\limits_{i = 1,j = 1}^{N_c} {d_{i,j}^s} } \right),
\end{equation}
where $s^{*}$ is the selected style image. 
In practice, we choose $N_c=3$, as involving too many color themes enforces the style of the generated video taking after that of the style image. 
For untargeted attacks, the style image is selected based on Equation~\ref{equ:color_sum}.

\subsubsection{Stylized Boundary Search}
To reduce the number of queries in targeted attacks, StyleFool conducts an offline search of style images that help approximate classification boundaries prior to sending queries to the target model.
The fitness of each style image is evaluated by its prediction confidence in the target class. Due to the fact that style transfer minimizes the style loss between a source video and the style image, the stylized video is similar to the style image in their feature representations. When there exists a transferability between the feature extraction network and the target model, the video stylized by a style image in the target class is likely to be closer to the decision boundary of the target class.
Therefore, adversarial videos can be crafted under fewer optimization steps after the stylization.

As presented in Equation~\ref{equ:score}, to select the best style image from the target class, we first convert each candidate image $s \in {S_t}$ in the style set $S_t$ into the video $x_{tile}$ by tiling $s$ (\ie using $s$ as the video frames repeatedly) according to the input frame requirement of the video classifier. The current top-1 label and the corresponding confidence score of $f(\cdot)$ on $x_{tile}$ are obtained.
If the top-1 label of $x_{tile}$ is the target class $y_t$, we record its confidence score as $p(y_t|x_{tile})$. Otherwise, the confidence score of $s$ will be set as 0. 
\begin{equation}\label{equ:score}
scor{e^t_s} = \left\{ {\begin{array}{*{20}{l}}
{p\left( {{y_t}\left| x_{tile} \right.} \right),~~if~f\left( x_{tile} \right) = {y_t}},\\
{0,~~otherwise.}
\end{array}} \right.
\end{equation}
In targeted attacks, we consider both the color theme proximity and $scor{e^t_s}$ of the style image at the same time. The principle of style image selection is updated as Equation~\ref{equ:targeted_criteria} in targeted attacks.
\begin{equation}\label{equ:targeted_criteria}
{s^*} = \mathop {\arg \min }\limits_{s \in {S_t}} \left( {\sum\limits_{i = 1,j = 1}^{N_c} {d_{i,j}^s}  + \mu \left( {1 - score_s^t} \right)} \right),
\end{equation}
where $\mu $ is a weight coefficient.

Note that the color themes of all images and target class confidence of all targets can be calculated and stored before style transfer, which can save both query number and time cost when performing a batch video attack. In contrast, one of the state-of-the-art attacks, H-Opt~\cite{wei2020heuristic}, needs a significant amount of queries before the adversarial generation stage.

\subsection{Style Transfer}
With the potential defects of video style transfer in mind (such as the temporal inconsistency~\cite{huang2017real} and the flicker~\cite{gao2018reconet}), during the video style transfer stage, both stylized characteristics and temporal consistency should be considered~\cite{ruder2016artistic}. The total loss of a video transfer can be expressed as
\begin{small}
\begin{equation}
\label{equation:total_loss}
\begin{array}{*{20}{l}}
{{L_{total}} = \sum\limits_{i = 1}^T {\left( {\alpha {L_{content}}\left( {{x_i},x_i^s} \right) + \beta {L_{style}}\left( {{x_i^s},s} \right) + \gamma {L_{tv}}\left( {{x_i^s}} \right)} \right)} }\\
{ + \lambda \sum\limits_{i = 1}^{T - 1} {\left( {{L_{temporal}}\left( {{x_i^s},{x_{i+1}^s}} \right)} \right)}, }
\end{array}
\end{equation}
\end{small}%
where $s$ is the style image (to simplify the notation, we use $s$ to present the style image in the rest of this paper); $x_i$ is the $i$-th frame of the initial clean video; and $x_i^s$ is the $i$-th frame of the stylized video ${x^s}$. ${{L_{content}}}$, ${{L_{style}}}$ and ${{L_{tv}}}$ represent the content loss, style loss, and total variation regularizer loss, respectively. $\alpha$, $\beta$, $\gamma$, and $\lambda$ are weight coefficients, and ${{L_{temporal}}}$ denotes the temporal loss between two consecutive frames. 

\noindent \textbf{Content loss.~}
Content loss is presented as the normalized Euclidean distance between the initial video $x$ and the stylized video ${x^s}$ across all layers of the feature map in VGG-19~\cite{simonyan2014very}, which indicates the dissimilarity between the initial video and the stylized video in their high-level representations. Minimizing such content loss encourages $x$ and ${x^s}$ to share semantic similarities in their content.
The content loss can be expressed as follows:
\begin{equation}
{L_{content}}\left( {{x_i},x_i^s} \right) = \sum\limits_k {\frac{1}{{{H_k}{W_k}{C_k}}}\left\| {{\vartheta _k}\left( {{x_i}} \right) - {\vartheta _k}\left( {x_i^s} \right)} \right\|_2^2},
\end{equation}
where ${H_k}$, ${W_k}$, and ${C_k}$ represent the height, width and channel number of the $k^{th}$ layer of the feature map, respectively. ${\vartheta _k}\left(  \cdot  \right)$ represents the feature map in the $k^{th}$ layer. 

\noindent \textbf{Style loss.~}
In order to reduce the difference between the style image and the stylized video, a style loss is introduced as the sum of the style errors across all layers in the style transfer network, which could be expressed as follows:
\begin{equation}
{L_{style}}\left( {{x_i^s},s} \right) = \sum\limits_k {\frac{1}{{C_k^2}}\left\| {{G_k}\left( s \right) - {G_k}\left( {x_i^s} \right)} \right\|_2^2},
\end{equation}
where ${G_k}\left(  \cdot  \right)$ represents the Gram matrix in the $k^{th}$ layer~\cite{justin2016perceptual}. 
The content loss and the style loss can move the input video towards or even across the decision boundary and alleviate the attack difficulty (\eg increase the attack success rate under query limit and reduce the query number) in black-box attacks, which are validated by the experiments in Section~\ref{sec:experimental_results}. 

\noindent \textbf{Total variation loss.~}
A total variance loss can be regarded as a regular term in the loss function, which eliminates noise or shadow and improves the smoothness of each stylized image. 
Given a frame $x_i^s$, a sum of pairwise variance is computed over all pixels in $x_i^s$ as follows:
\begin{equation}
\begin{array}{l}
{L_{tv}}\left( {{x_i^s}} \right) = \sum\limits_{u,v} {\left( {{{\left\| {x_i^s\left( {u,v} \right) - x_i^s\left( {u + 1,v} \right)} \right\|}^2}} \right.} \\
\left. { + {{\left\| {x_i^s\left( {u,v} \right) - x_i^s\left( {u,v + 1} \right)} \right\|}^2}} \right),
\end{array}
\end{equation}
where $x_i^s\left( {u,v} \right)$ represents the pixel value at the coordinate $\left( {u,v} \right)$ in $x_i^s$.

\noindent \textbf{Temporal loss.~}
The temporal loss is used to improve the consistency between two consecutive frames, which can be expressed as
\begin{equation}
{L_{temporal}}\left( {{x_i^s},{x_{i+1}^s}} \right) = \frac{1}{{HWC}}{O_{i + 1}}{\left\| {x_{i + 1}^s - \mathcal W\left( {x_i^s} \right)} \right\|^2},
\end{equation}
where ${\mathcal W}\left(  \cdot  \right)$ represents a warp function~\cite{ruder2016artistic} that outputs the warped frame generated from the input frame using pre-computed optical flow; $O_{i + 1}$ represents an occlusion mask matrix of the $\left( i+1 \right)^{th}$ frame in optical flow.

Thus, the stylized video $x^s$ is computed by minimizing the total loss in Equation~\ref{equation:total_loss}, which has the characteristics of smoothness and high temporal consistency in parallel. Such temporal consistency in the stylized video can be retained in generated adversarial samples.
Therefore, the introduction of temporal loss provides a guarantee for StyleFool to bypass the adversarial defense mechanism based on time consistency, \eg AdvIT~\cite{xiao2019advit}.

\noindent \textbf{Style loss and Maximum Mean Discrepancy.~}
Another natural method would be optimizing the videos via a surrogate classifier prior to querying the target classifier. 
However, we discover that style transfer is superior to surrogate classifiers. It is known that minimizing the style loss is equivalent to minimizing the Polynomial kernel Maximum Mean Discrepancy (PMMD) between two sets ${V_k}\left( x^s_i\right)$ and ${V_k}\left( s\right)$ of feature vectors~\cite{li2017demystifying}.
Let $M_k=H_k\times W_k$, ${V_k}\left( \cdot\right):= \{{\vartheta _k}\left( \cdot\right)_p|{\vartheta _k}\left( \cdot\right)_p\in\R^{C_k}\}^{M_k}_{p=1}$.
${\vartheta _k}\left( \cdot\right)_p$ is the $p$-th vector in the $C_k\times M_k$ feature map.
Minimizing PMMD aligns the feature distributions of $x_i^s$ and $s$. 
Such alignment may intrinsically increase the transferability of the optimized $x^s$ on different networks since the frames of $x^s$ are distributionally similar to $s$ in the feature space.
On the other hand, a surrogate classifier only minimizes the difference between the prediction of the optimized video and the target label without deliberately modifying the intrinsic feature distribution.
Henceforth, the transferability of the surrogate-optimized video is susceptible to changes in the classifier parameters or architectures.
As an evidence, the query numbers of attacks based on stylized videos are significantly less than those based on surrogate-optimized videos (check Table~\ref{tab_surrogate_performance} in Appendix~\ref{appendix_exp_results}).

\subsection{Adversarial Sample Generation}
In the black-box attacks, since the attacker cannot have access to the model structure and parameters, it is impossible to establish the loss function and optimize it with the gradient information. In order to optimize the loss function in the black-box setting, Natural Evolution Strategy (NES), one of the most efficient gradient estimation methods~\cite{ilyas2018black} is introduced to estimate the gradients. 

For an adversarial loss function $L\left( \theta  \right)$ under a search distribution $\pi \left( {\theta \left| x \right.} \right)$, the gradient estimates ${\hat g}$ can be expressed as
$\hat g = {\nabla _x}{{\rm{E}}_{\pi \left( {\theta \left| x \right.} \right)}}\left[ {L\left( \theta  \right)} \right]$.
By applying a log-trick inside the integral form of the above expectation, it yields
\begin{align}
    {\nabla _x}{{\rm{E}}_{\pi \left( {\theta \left| x \right.} \right)}}\left[ {L\left( \theta  \right)} \right]
    &= {\nabla _x}\int{L(\theta)}{\pi \left( {\theta \left| x \right.} \right)}{d}{\theta}\\\nonumber
    &= \int{L(\theta)}{\frac{\pi \left( {\theta \left| x \right.} \right)}{\pi \left( {\theta \left| x \right.} \right)}}{\nabla _x}{\pi \left( {\theta \left| x \right.} \right)}d{\theta}\\\nonumber
    &= {{\rm{E}}_{\pi \left( {\theta \left| x \right.} \right)}}
    \left[ {L\left( \theta  \right){\nabla _x}\log \pi \left( {\theta \left| x \right.} \right)} \right].
\end{align}
In this way, it only requires to query the target model for the value of $L(\theta)$. To reduce the variance of the gradient estimates $\hat{g}$, we apply antithetic sampling to retrieve $n_g$ points around $x$ under a Gaussian search distribution, \ie half of the noise samples are generated by ${\theta _i} = x + \sigma {\delta _i}$ for $i \in \left\{ {1,...,\frac{n_g}{2}} \right\}$, where $\sigma $ is the standard deviation of the noise, ${\delta _i}  \sim N\left( {0,I} \right)$, $I$ is an identity matrix. 
Then we invert the first half of the noise to obtain another half of the noise, \ie ${\delta _i} =  - {\delta _{n_g + 1 - i}}$, for $i \in \left\{ {\frac{n_g}{2} + 1,...,n_g} \right\}$. 
The gradient estimates can be approximated as follows:
\begin{align}
\begin{array}{l}
\hat g \approx \frac{1}{{n_g\sigma }}\sum\limits_{i = 1}^{n_g/2} {{\delta _i}L\left( {x + \sigma {\delta _i}} \right)} \\
 + \frac{1}{{n_g\sigma }}\sum\limits_{i = n_g/2 + 1}^{n_g} {\left( { - {\delta _{n_g + 1 - i}}} \right)L\left( {x - \sigma {\delta _{n_g + 1 - i}}} \right)}.
\end{array}
\end{align}
Since antithetic sampling reduces the variance of the gradient estimates, the above equation can be regarded as a variance-reduced approximation of Stein's Lemma~\cite{stein1981estimation}.

For black-box targeted attacks, only the top-1 label and its confidence score can be accessed, making it difficult to find the appropriate gradient descent direction~\cite{ilyas2018black, brendel2017decision}. Therefore, the attack begins with an instance $x_{adv}$ selected from the target class $y_t$ and is optimized by Projected Gradient Decent (PGD)~\cite{madry2017towards} with the gradient estimate.
Since StyleFool has selected the most appropriate style image from the target class in the style transfer stage, the stylized video has carried a large number of features contained in the style image. In other words, higher similarity indicates closer distance in high-dimensional space, which is beneficial to adversarial attacks. Therefore, we choose the video $x_{vs}$ where the style image is located as the initial target class instance. 
In each PGD iteration, backtracking line search is introduced to find the minimal possible perturbation size which ensures that the target class $y_t$ remains the top-1 label. 
Attack succeeds when {the initial perturbation}
$\varepsilon$ (initialized as 1) goes down and reaches the perturbation threshold $\varepsilon_{adv}$. 
This ensures the adversarial video generated from the stylized video is difficult to be distinguished by human eyes. In order to improve the attack efficiency, untargeted attacks directly take the stylized video $x^s$ as the initial value of ${x_{adv}}$. 

The whole process of StyleFool proposed in this paper is shown in Algorithm~\ref{alg:style_attack}. In this algorithm, $y_t$ is $None$ in the untargeted attack. $\init$ returns the initial style set which contains a large number of style images, $\colorprox$ outputs the color theme proximity, $\score$ gives out a target class confidence score of a style image, $\transfer$ delivers the style transfer, $\tovideo$ finds the video to which the style image belongs, and $\BPGD\left(  \cdot  \right)$ is an extension to PGD with backtracking line search and learning rate adjustment, the details of which are given in the work done by Ilyas~\etal~\cite{ilyas2018black}. 

\noindent
\begin{algorithm}[t]\small
\footnotesize
\caption{StyleFool.}\label{alg:style_attack}
\KwIn{Black-box classifier $f$, input video ${x_0}$, input label $y_0$, target class ${y_t}$, style set $S$, perturbation threshold ${\varepsilon_{adv}}$, initial perturbation $\varepsilon$, total loss $L_{total}$, adversarial loss $L$, step size $\eta$.}
\KwOut{Adversarial video ${x_{adv}}$.}
$S \gets \init()$\;
\For {$k \gets 1$ to $\len(S)$}{
$c[k] \gets \colorprox(x_0,S[k])$\;
\If{$y_t \ne \none$}{
$c[k] \gets c[k] + \score(f,S[k],y_t)$\;
}
}
$idx \gets \argmin(c)$\;
$s^{*} \gets S[idx]$\;
${x^s} \gets \transfer(x_0,s^{*},L_{total})$\;
\eIf{$y_t == \none$}{
$x_{adv} \gets x^s$\; 
\While {$f(x_{adv}) == y_0$}{
$\hat g \gets \NES(x_{adv},L)$\;
$x_{adv} \gets \PGD(x_{adv},x^s,\hat g,\eta,\varepsilon_{adv})$\;
}
}
{
${x_{adv}} \leftarrow \tovideo(s^{*})$\; 
\While {${\varepsilon} > \varepsilon_{adv}$ or $f(x_{adv}) \ne y_t$}{
$\hat g \leftarrow \NES(x_{adv},L)$\;
$x_{adv},\eta,\varepsilon \leftarrow \BPGD(x_{adv},x^s,\hat g,\eta,\varepsilon)$\;
}
}
\end{algorithm}
\vspace{-3mm}

\subsection{StyleFool Recap}
In untargeted attack scenarios, for an input video, the style is selected based on the color theme proximity (line~3); while in targeted attack scenarios, the attacker will first randomly select a target label, then choose a style image in the target label according to the color theme proximity (line 3) and the target class confidence (line 5). Then, the input video is transferred to a stylized video according to the selected style (lines 6 to 8). Finally, NES and PGD are used to generate adversarial videos under black-box setting (lines 9 to 18). The required adversarial sample ${x_{adv}}$ is generated at the end. 

\section{Evaluation}\label{sec:evaluation}
In this section, we carry out comprehensive experiments to evaluate the performance and indistinguishability of StyleFool. Moreover, we provide possible variants of StyleFool, verify the importance of style selection in the ablation study, and reveal the advantage of style transfer through a quantitative evaluation.

\subsection{Experiment Setup}

\noindent \textbf{Datasets.~} 
We employ two widely used datasets, UCF-101~\cite{soomro2012ucf101} and HMDB-51~\cite{kuehne2011hmdb} to validate the attack performance of the proposed StyleFool.
\begin{itemize}[leftmargin=*]
    \item \textbf{UCF-101} is an action recognition dataset collected from YouTube, containing 13,320 video samples with 101 action classes, \eg archery, haircut, and punch.
    \item \textbf{HMDB-51} is a collection of realistic videos from various sources, including movies and web videos, containing 6,849 video samples with 51 action classes, \eg sword, climb, and golf.
\end{itemize}

We randomly select 70\% of the videos in the datasets as the style set, and then the style image is selected from the style set using the criterion in Section~\ref{sec:style selection}. The remaining 30\% of the videos are used for adversarial sample generation. 

\noindent \textbf{Target video classifiers and evaluation metrics.}
We involve C3D~\cite{tran2015learning} and I3D~\cite{carreira2017quo} as our targets. 
The two models utilize different strategies and achieve the state-of-the-art video classification performance.
C3D learns both spatial and temporal features of input videos using 3D convolution, while I3D utilizes optical flow to build the relationship between two adjacent frames. Since C3D requires videos with 16 frames as input, we separate all videos into 16-frame snippets. The classification performance of C3D and I3D is shown in Table~\ref{tab:classification_accuracy}.
We define the following metrics in the evaluation. 

\begin{itemize}[leftmargin=*]

\item \textbf{Attack Success Rate (ASR)}: the ratio of adversarial videos that successfully mislead the classifier. Note that any attack exceeding the query limit will be considered as failed. 

\item \textbf{Minimal Queries (minQ), Maximal Queries (maxQ), and Average Queries (AQ)}: the minimal, maximal, and average numbers of queries to succeed in an attack. For fair comparison, in StyleFool, the number of queries during style selection is also counted, although this proportion is quite small (as analyzed in Section~\ref{sec:experimental_results}).

\item \textbf{Indistinguishability}: the \textit{naturalness} (\ie the generated adversarial videos are expected to preserve the semantic information of original videos and appear natural to human subjects) and \textit{realness} (\ie the style transferred samples should retain the video quality of experience and mislead human subjects into thinking they are non-artificial videos) of videos. 
Additionally, the SSIM~\cite{wang2004ssim}, PSNR, and FID~\cite{dowson1982fid} are used to statistically measure the indistinguishability, which are widely used to evaluate image quality. We extend them to video domain by averaging the per-frame metrics over the entire video. The equation of the PSNR is slightly modified to $PSNR = 10{\log _{10}}\frac{{{{255}^2}}}{{\left( {MSE + {\alpha _m}} \right)}}$ to avoid images with no difference, where $MSE$ is the mean square error between two images, ${\alpha _m} = {10^{ - 5}}$.

\end{itemize}

\noindent \textbf{Benchmarking attack frameworks.} 
In order to compare $\ell_p$-norm attacks and unrestricted attacks, we extend the NES-based $\ell_p$-norm attack on images (denoted by the first author's name as Ilyas~\cite{ilyas2018black}) and apply it onto video samples. Additionally, we compare our framework with two state-of-the-art one-on-one black-box attack frameworks, V-BAD~\cite{jiang2019black} and H-Opt (Heuristic Opt-based Black-box Attack)~\cite{wei2020heuristic}. 
V-BAD reduces the computational complexity by splitting tentative perturbations into several patches and rectifying the original gradient direction.
Based on the Opt attack~\cite{cheng2018query} in images, H-Opt uses Zeroth Order Optimization (ZOO)~\cite{chen2017zoo} to update the video. H-Opt considers adding sparse perturbations by key frames extraction and saliency detection. 

Since C-DUP~\cite{li2019stealthy} and U3D~\cite{xie2022universal} are universal perturbations only suitable for untargeted attacks, it is unreasonable and unfair to compare them with one-on-one attack frameworks (\eg V-BAD, H-Opt, and StyleFool) due to different attack purposes.
Therefore, we only consider Ilyas~\cite{ilyas2018black}, V-BAD~\cite{jiang2019black}, and H-Opt~\cite{wei2020heuristic} as our benchmarks in the experiments. 
Considering that existing black-box attacks~\cite{papernot2017practical,chen2017zoo,ilyas2018black} require about $10^4$ queries on CIFAR-10, and about $10^5$ queries on ImageNet to succeed, we set a reasonable query limit as $3 \times {10^5}$, which is similar to the setting of the research by Jiang~\etal~\cite{jiang2019black}. The perturbation threshold $\varepsilon_{adv}$ is set as 0.05. 
We provide other configurations of StyleFool in Appendix~\ref{appendix_configuration}.

\begin{table}[t]  
\centering
\caption{Video classification accuracy of C3D and I3D.}
\label{tab:classification_accuracy}
\vspace{1mm}
\resizebox{0.4\linewidth}{!}{
\begin{tabular}{ccc}
\toprule
\multirow{2}{*}[-0.5ex]{\textbf{Model}} & \multicolumn{2}{c}{\textbf{Datasets}}\\
\cmidrule(r){2-3}
 & \textbf{UCF-101} & \textbf{HMDB-51} \\
\midrule
C3D & 84.9\% & 67.1\% \\
I3D & 87.6\% & 62.5\% \\
\bottomrule
\end{tabular}
}
\end{table}

\begin{table*}[t]  
\centering
\caption{{Attack performance comparison.}}
\label{tab:attack_performance}
\resizebox{0.9\linewidth}{!}{
\begin{tabular}{ccrrrrrrrrrrrrrrrr}
\toprule
\multirow{2}{*}[-0.5ex]{\textbf{Model}} & 
\multirow{2}{*}[-0.5ex]{\textbf{Attack}} & \multicolumn{4}{c}{\textbf{UCF-101 (Targeted)}} & \multicolumn{4}{c}{\textbf{HMDB-51 (Targeted)}}  & \multicolumn{4}{c}{\textbf{UCF-101 (Untargeted)}} & \multicolumn{4}{c}{\textbf{HMDB-51 (Untargeted)}}\\
\cmidrule(r){3-6}\cmidrule(r){7-10}\cmidrule(r){11-14}\cmidrule(r){15-18}
& & \footnotesize{\textbf{ASR}} & \footnotesize{\textbf{minQ}} & \footnotesize{\textbf{maxQ}} & \footnotesize{\textbf{AQ}} &  \footnotesize{\textbf{ASR}} & \footnotesize{\textbf{minQ}} & \footnotesize{\textbf{maxQ}} & \footnotesize{\textbf{AQ}} &  \footnotesize{\textbf{ASR}} & \footnotesize{\textbf{minQ}} & \footnotesize{\textbf{maxQ}} & \footnotesize{\textbf{AQ}} &  \footnotesize{\textbf{ASR}} & \footnotesize{\textbf{minQ}} & \footnotesize{\textbf{maxQ}} & \footnotesize{\textbf{AQ}} \\
\midrule
\multirow{4}{*}{C3D} & Ilyas~\cite{ilyas2018black} & 62 & 43,112 & $>$300,000 & $>$209,847 & 96 & 4,302 & $>$300,000 & $>$96,563 & \textbf{100} & 50 & 58,948 & 24,202 & \textbf{100} & 50 & 47,727 & 6,631 \\

& V-BAD~\cite{jiang2019black} & 85 & 1,269 & $>$300,000 & $>$117,053 & 98 & 517 & $>$300,000 & $>$56,767 & \textbf{100} & 50 & 150,725 & 12,646 & \textbf{100} & 50 & 42,974 & 4,083 \\
& H-Opt~\cite{wei2020heuristic} & 60 & 16,989 & $>$300,000 & $>$228,867 & 70 & 10,010 & $>$300,000 & $>$195,242 & \textbf{100} & 1,020 & 97,929 & 19,131 & \textbf{100} & 1,948 & 42,183 & 14,066\\
& StyleFool & \textbf{100} & \textbf{1,284} & \textbf{248,131} & \textbf{65,066} & \textbf{100} & \textbf{101} & \textbf{95,897} & \textbf{35,232} & \textbf{100} & \textbf{1} & \textbf{19,013} & \textbf{3,811} & \textbf{100} & \textbf{1} & \textbf{9,948} & \textbf{1,526} \\
\midrule
\multirow{4}{*}{I3D} & Ilyas~\cite{ilyas2018black} & 95 & 10,875 & $>$300,000 & $>$91,495 & \textbf{100} & 4,344 & 246,149 & 64,047 & \textbf{100} & 1,177 & 139,749 & 25,187 & \textbf{100} & 197 & 33,811 & 7,234\\

& V-BAD~\cite{jiang2019black} & 99 & 101 & $>$300,000 & $>$56,680 & \textbf{100} & 521 & 211,138 & 42,508 & \textbf{100} & 148 & 123,236 & 10,437 & \textbf{100} & 50 & 30,822 & 3,447 \\
& H-Opt~\cite{wei2020heuristic} & 31 & 8,583 & $>$300,000 & $>$260,463 & 53 & 6,892 & $>$300,000 & $>$212,256 & \textbf{100} & 3,165 & 156,624 &  47,008 & \textbf{100} & 1,386 & 164,580 & 37,897 \\
& StyleFool & \textbf{100} & \textbf{101} & \textbf{119,991} & \textbf{30,876} &  \textbf{100} & \textbf{101} & \textbf{80,319} & \textbf{23,556} & \textbf{100} & \textbf{1} & \textbf{30,724} & \textbf{6,643} & \textbf{100} & \textbf{1} & \textbf{6,665} & \textbf{2,013}\\
\bottomrule
\end{tabular}
}
\end{table*}

\subsection{Experimental Results} \label{sec:experimental_results}

\noindent \textbf{Attack performance.~} 
In order to quantitatively compare the attack performance of StyleFool with benchmark attacks, we randomly select 100 videos from each dataset as the initial videos and mount both targeted and untargeted attacks on datasets UCF-101 and HMDB-51. Table~\ref{tab:attack_performance} shows the attack performance of the four frameworks, \ie Ilyas, V-BAD, H-Opt, and StyleFool. 
Due to the restrictions on the number of queries (at most $3 \times 10^5$ queries) and partial information (the adversary can only access the top-1 label and its confidence score), the attack success rates of Ilyas, V-BAD and H-Opt cannot always achieve 100\%, while StyleFool is stable at 100\% for all scenarios; for some video samples, Ilyas, V-BAD and H-Opt cannot complete the attack within the query limit. Note that black-box query-based attacks will always succeed if there is no query limit.
However, in StyleFool, even input videos that are far from the target class decision boundary will not cause the query number to exceed the upper limit, as style transfer will push the adversarial samples towards the boundary and thus fewer queries are ensured at the same time. 

According to the average queries results, compared with the benchmark frameworks, StyleFool requires the least number of queries to succeed in an attack. Specifically, the query number of StyleFool is at least 69\% and 44\% and 72\% less than Ilyas, V-BAD and H-Opt, respectively, in targeted attacks on C3D using the UCF-101 dataset. When the attacked model is I3D, in untargeted attacks conducted on the HMDB-51 dataset, the average queries of StyleFool are far fewer than those of H-Opt (2,013 vs. 37,897). On average, StyleFool reduces 65\% (77\%), 43\% (53\%), and 83\% (87\%) of queries to succeed in the targeted (untargeted) attack, compared with Ilyas, V-BAD and H-Opt, respectively. 
Overall, StyleFool requires the least queries to succeed in an attack. Specifically, in cases where the number of queries is 1 (\eg the minQs in untargeted attacks), the stylized video is already adversarial, and thus no more queries are needed.
As a result, StyleFool demonstrates the efficiency of attacking with $(2.4 \sim 6.5) \times {10^4}$ queries for targeted attacks and $( 1.5 \sim 6.6) \times {10^3}$ queries for untargeted attacks. The experimental results demonstrate that StyleFool is more generalized and efficient, compared with the state-of-the-art approaches. Figure~\ref{Fig:visualization_StyleFool_simplify} in Appendix~\ref{appendix_exp_results} shows the adversarial video samples generated by StyleFool in targeted and untargeted attacks. The styles of original videos have been transferred and the goal of adversarial attack has also been achieved. 

\noindent \textbf{Analysis.~}
The reason why Ilyas requires more queries is that in the high-dimensional feature space, it is more difficult to pull the video from the target class back into the $\ell_p$ ball of the original video, and randomly selecting initial target class video makes the attack difficulty uncertain. Note that Ilyas is a special case of StyleFool without style transfer. 
Since H-Opt needs many queries in finding the initial attack direction and calculating every decision boundary distance using fine-grained search and binary search, the average queries remain high as a whole. When the query limit is set, the attack fails when the number of queries is greater than $3 \times {10^5}$. Therefore, the attack success rate of H-Opt is significantly lower than that of the other two frameworks. 
Though H-Opt considers sparse perturbations temporally and spatially, such a weak advantage cannot be offset by the high cost of query number which is one order of magnitude higher than other frameworks, since query number is the most important index of black-box attack. 
V-BAD improves more than H-Opt. However, its performance is still not as good as StyleFool. The style transfer in StyleFool moves videos closer to the decision boundary, which is beneficial to adversarial attacks, resulting in fewer queries. The average queries of StyleFool are 40-70\% fewer than that of V-BAD. 

Note that in the style selection stage, we need to query the classifier for targeted attacks to obtain the target class confidence of all style images, but the number of queries is extremely smaller in comparison with the adversarial sample generation stage. When attacking a C3D model on the UCF-101 dataset, H-Opt spends about 15\% of queries (approximately $3 \times {10^4}$) on searching an initial target class video and extracting key frames, while StyleFool spends only 0.45\% of queries (approximately  $3 \times {10^2}$) during the style selection stage. 

\subsection{Indistinguishability} 

\begin{figure}[t]
    \centering
    \includegraphics[width=0.8\linewidth]{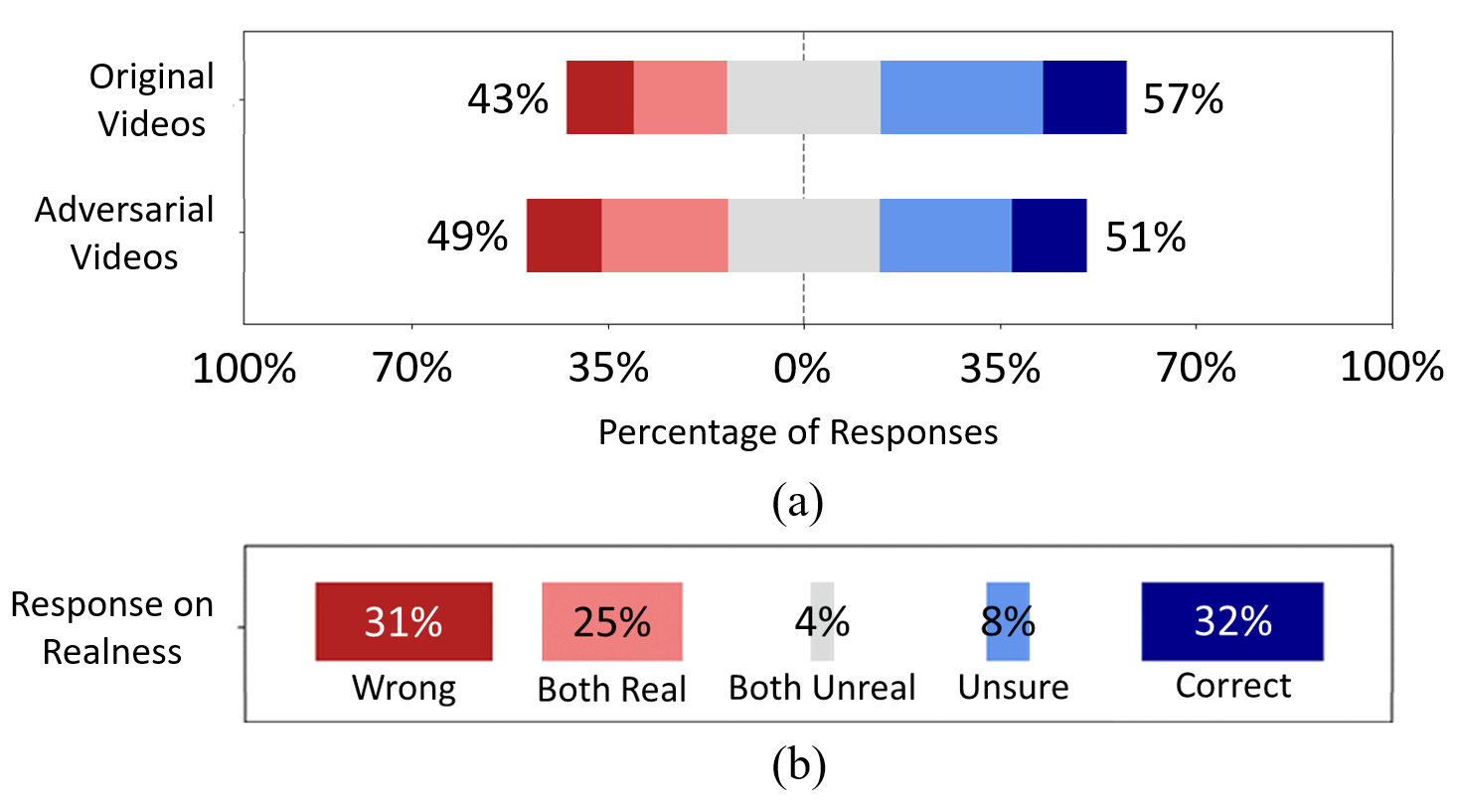}
    \caption{Results of the user study on indistinguishability: (a) Participants' responses in the naturalness test. \firebrick~: very unnatural, \darkblue~: very natural; (b) Participants' responses in the realness test.}
    \label{fig:user study}
\end{figure} 

In order to verify the indistinguishability of StyleFool, we conducted a user study to evaluate the naturalness and realness of adversarial samples (randomly selected from the samples generated from the experiment in Section~\ref{sec:experimental_results}).
We designed a web questionnaire where both adversarial videos and clean videos are displayed and recruited {\nParticipants} anonymous participants in an online survey via Amazon Mechanical Turk (AMT)~\cite{amazon_turk}. The Human Research Ethics Committee of lead author’s affiliation determined that the study was
exempt from further human subjects review, and we followed best practice for ethical human subjects survey research, \eg all questions were optional and we did not collect unnecessary personal information. Participant consented for their answers to be used for academic research. All participants are over 18 years old and are able to complete the survey in English.
Appendix~\ref{appendix_user_study} provides the survey protocol and more detailed settings (\eg the demographics and the payment). 

\noindent \textbf{Survey results and data analysis.~}
The average completion time of participants in the naturalness test and the realness test was 59.4 minutes and 62.5 minutes respectively. According to the participants' responses about naturalness shown in Figure~\ref{fig:user study}(a), we find that participants could not distinguish between clean videos and adversarial videos very well.
51\% of participants gave positive naturalness responses to adversarial videos, while 57\% of participants thought the original clean videos were natural -- only 6\% higher. It can be concluded that most participants did not give a high confidence score of naturalness to clean videos (15\% of participants voted ``very natural'' for clean videos), nor did they give a low confidence of naturalness to the adversarial video (13\% of participants voted ``very unnatural'' on adversarial samples).
Our variables are ordinal and the responses to each question are not expected to be normally distributed. Therefore, we conduct a Mann-Whitney U-test~\cite{mann1947on} for statistical significance testing. Concretely, we calculate the $p$-value of the aggregated Likert scale responses on original videos and adversarial videos with the null hypothesis $H_0$: there is no significant difference between the naturalness of original videos and adversarial videos. The U-test result ($p=0.07>0.05$) suggests that we cannot reject the null hypothesis, indicating a high degree of sensory comfort is maintained by adversarial videos generated by StyleFool.

Figure~\ref{fig:user study}(b) shows the distribution of participants' responses in the realness test. Only 32\% of the participants correctly distinguished adversarial videos from clean videos, and 31\% of the participants provided totally incorrect answers. Considering that the accurate rate is much lower than random guessing, we can conclude that the adversarial samples could not be identified by participants. Additionally, 25\% of participants said both videos were real, which indicates that adversarial videos looked as realistic as clean videos for them. 4\% of the participants thought that both videos were unreal, and 8\% got trapped in judgment difficulty. The accuracy of participants' judgment is 50.5\% with 50.4\% precision and 62.0\% recall, which indicates that the participants could not distinguish the adversarial videos well.

\noindent {\textbf{Quantitative analysis.~}}
Since StyleFool is an attack with unrestricted perturbations, we argue that it could be unfair to compare unrestricted attacks and restricted attacks by video quality metrics. However, we still would like to quantitatively evaluate the indistinguishability performance of StyleFool. We report the SSIM, PSNR, and FID between original videos and adversarial videos, shown as ``StyleFool-ori-adv'' in Table~\ref{tab_video_quality_performance} in Appendix~\ref{appendix_exp_results}, which is expected to be lower than existing restricted methods (such as V-BAD and H-Opt), but still good enough to keep stealthy. The changes in styles and textures are bound to cause differences in those three metrics on video quality, but the semantic information is not influenced (see Figure~\ref{Fig:visualization_StyleFool_simplify} in Appendix~\ref{appendix_exp_results}). We further evaluate the indistinguishability between stylized videos and adversarial videos (\ie the ``StyleFool-sty-adv'' in Table~\ref{tab_video_quality_performance}). As we particularly restrict the perturbations between stylized videos and the adversarial videos, the ``StyleFool-sty-adv'' shows the best indistinguishability performance among all benchmark methods, due to less modification in pixels. 

\subsection{Ablation Study}\label{sec:ablation} 

\begin{table*}[t]  
\centering
\caption{StyleFool performance with different $\varepsilon_{adv}$.}
\label{tab_ablation_results_epsilon}
\resizebox{0.8\linewidth}{!}{
\begin{tabular}{ccrrrrrrrrrrrrrrrr}
\toprule
\multirow{2}{*}[-0.5ex]{\textbf{Model}} & 
\multirow{2}{*}[-0.5ex]{\textbf{$\varepsilon_{adv}$}} & \multicolumn{4}{c}{\textbf{UCF-101 (Targeted)}} & \multicolumn{4}{c}{\textbf{HMDB-51 (Targeted)}}  & \multicolumn{4}{c}{\textbf{UCF-101 (Untargeted)}} & \multicolumn{4}{c}{\textbf{HMDB-51 (Untargeted)}}\\
\cmidrule(r){3-6}\cmidrule(r){7-10}\cmidrule(r){11-14}\cmidrule(r){15-18}
& & \footnotesize{\textbf{ASR}} & \footnotesize{\textbf{minQ}} & \footnotesize{\textbf{maxQ}} & \footnotesize{\textbf{AQ}} &  \footnotesize{\textbf{ASR}} & \footnotesize{\textbf{minQ}} & \footnotesize{\textbf{maxQ}} & \footnotesize{\textbf{AQ}} &  \footnotesize{\textbf{ASR}} & \footnotesize{\textbf{minQ}} & \footnotesize{\textbf{maxQ}} & \footnotesize{\textbf{AQ}} &  \footnotesize{\textbf{ASR}} & \footnotesize{\textbf{minQ}} & \footnotesize{\textbf{maxQ}} & \footnotesize{\textbf{AQ}} \\
\midrule
\multirow{4}{*}{C3D} 
& 0.05 & 100 & 1,284 & 248,131 & 65,066 & 100 & 101 & 95,897 & 35,232 & 100 & 1 & 19,013 & 3,811 & 100 & 1 & 9,948 & 1,526 \\
& 0.10 & 100 & 101 & 127,432 & 25,054 & 100 & 101 & 62,546 & 15,240 & 100 & 1 & 7,204 & 1,348 & 100 & 1 & 2,842 & 620 \\
& 0.15 & 100 & 101 & 83,134 & 12,297 & 100 & 101 & 33,268 & 6,490 & 100 & 1 & 5,538 & 1,155 & 100 & 1 & 2,402 & 457 \\
& 0.20 & 100 & 101 & 57,223 & 6,553 & 100 & 101 & 16,227 & 2,202 & 100 & 1 & 3,803 & 813 & 100 & 1 & 2,059 & 415 \\
\midrule
\multirow{4}{*}{I3D}
& 0.05 & 100 & 101 & 119,991 & 30,876 &  100 & 101 & 80,319 & 23,556 & 100 & 1 & 30,724 & 6,643 & 100 & 1 & 6,665 & 2,013\\
& 0.10 & 100 & 101 & 82,588 & 10,502 & 100 & 101 & 51,478 & 10,378 & 100 & 1 & 18,131 & 3,145 & 100 & 1 & 2,892 & 1,007 \\
& 0.15 & 100 & 101 & 63,918 & 5,050 & 100 & 101 & 29,953 & 4,653 & 100 & 1 & 16,220 & 2,802 & 100 & 1 & 2,304 & 927 \\
& 0.20 & 100 & 101 & 40,895 & 2,237 & 100 & 101 & 20,344 & 2,480 & 100 & 1 & 13,280 & 1,873 & 100 & 1 & 2,108 & 899 \\
\bottomrule
\end{tabular}
}
\end{table*}

\noindent \textbf{Possible variants of StyleFool.} 
Since StyleFool is an unrestricted attack, the perturbation threshold, $\varepsilon_{adv}$, could be varied, instead of strictly setting to 0.05 in the process of PGD. Properly increasing $\varepsilon_{adv}$ may not affect the visual sense. To determine the influence of larger perturbation, we set $\varepsilon_{adv}$ to 0.05, 0.10, 0.15 and 0.20, respectively, and conduct ablation experiments on two datasets with two models. Table~\ref{tab_ablation_results_epsilon} and Figure~\ref{Fig:visualization_StyleFool_epsilon} in Appendix~\ref{appendix_exp_results} show the attack performance and visualization of StyleFool under different $\varepsilon_{adv}$. The average queries are reduced sharply with the increase of $\varepsilon_{adv}$. However, since targeted attacks start with a video from the targeted label, a larger $\varepsilon_{adv}$ will lead to the superposition of two videos in the adversarial video (\eg $\varepsilon_{adv} = 0.20$ in targeted attacks shown in Figure~\ref{Fig:visualization_StyleFool_epsilon}(a)), although the average queries are reduced by an order of magnitude. We point out that there is a trade-off between the number of queries and visual perception.
While in untargeted attacks, a larger $\varepsilon_{adv}$ will not exert much impact on the visual perception of the adversarial video (\eg $\varepsilon_{adv} = 0.10, 0.15$, and $0.20$ in untargeted attacks shown in Figure~\ref{Fig:visualization_StyleFool_epsilon}(b)). In addition, StyleFool with a larger $\varepsilon_{adv}$ can largely reduce queries, \eg in untargeted attacks against C3D trained on the HMDB51 dataset, StyleFool with $\varepsilon_{adv} = 0.20$ reduces average queries by 73\%, compared with $\varepsilon_{adv} = 0.05$. To conclude, a larger perturbation threshold is practicable in StyleFool, but the trade-off between the number of queries and visual perception should also be considered, especially in targeted attack scenarios.

\begin{table}[t]  
\centering
\caption{Results of StyleFool in different scenarios: \one~randomly selecting styles; \two~only color theme proximity is considered (untargeted StyleFool); \three~only target class confidence is considered; and \four~both color theme and target class confidence are considered (targeted StyleFool).}
\label{tab_ablation_result_style}
\resizebox{\linewidth}{!}{
\begin{tabular}{lrrrrrrrr}
\toprule
\multirow{2}{*}[-1ex]{\textbf{Attack}} & \multicolumn{4}{c}{\textbf{UCF-101 (Targeted)}} & \multicolumn{4}{c}{\textbf{HMDB-51 (Targeted)}}\\
\cmidrule(r){2-5}\cmidrule(r){6-9}
& \footnotesize{\textbf{ASR}} & \footnotesize{\textbf{minQ}} & \footnotesize{\textbf{maxQ}} & \footnotesize{\textbf{AQ}} &  \footnotesize{\textbf{ASR}} & \footnotesize{\textbf{minQ}} & \footnotesize{\textbf{maxQ}} & \footnotesize{\textbf{AQ}} \\
\midrule
\textbf{Random} & \textbf{100} & \textbf{101} & 191,600 & 52,285  & \textbf{100} & \textbf{101} & 154,299 & 37,232 \\
\textbf{Color theme proximity} & \textbf{100} & 7,416 & 257,676 & 77,642 & \textbf{100} & 5202 & 149,232 & 49,227 \\
\textbf{Target class confidence} & \textbf{100} & \textbf{101} & 211,110 & 55,534 & \textbf{100} & \textbf{101} & 156,875 & 43,575 \\
\textbf{Both} & \textbf{100} & \textbf{101} & \textbf{106,713} & \textbf{44,096} & \textbf{100} & \textbf{101} & \textbf{68,850} & \textbf{31,894}\\
\midrule
\multirow{2}{*}[-1ex]{\textbf{Attack}} & \multicolumn{4}{c}{\textbf{UCF-101 (Untargeted)}} & \multicolumn{4}{c}{\textbf{HMDB-51 (Untargeted)}}\\
\cmidrule(r){2-5}\cmidrule(r){6-9}
&  \footnotesize{\textbf{ASR}} & \footnotesize{\textbf{minQ}} & \footnotesize{\textbf{maxQ}} & \footnotesize{\textbf{AQ}} &  \footnotesize{\textbf{ASR}} & \footnotesize{\textbf{minQ}} & \footnotesize{\textbf{maxQ}} & \footnotesize{\textbf{AQ}}\\
\midrule
\textbf{Random} & \textbf{100} & \textbf{1} & 18,416 & 2,837 & \textbf{100} & \textbf{1} & 11,712 & 1,236 \\
\textbf{Color theme proximity} & \textbf{100} & \textbf{1} & \textbf{13,563} & \textbf{2,228} & \textbf{100} & \textbf{1} & \textbf{4,068} & \textbf{937} \\
\bottomrule
\end{tabular}
}
\end{table}

\noindent \textbf{Contribution of style selection in StyleFool.} 
To further explore how color proximity and target class confidence contribute to the performance of StyleFool, we next conduct an ablation study.
Concretely, we carry out experiments in the following four different style selection scenarios: \one~randomly selecting styles; \two~only color theme proximity is considered (StyleFool in untargeted attacks); \three~only  target class confidence is considered; \four~both color theme proximity and target class confidence are considered (StyleFool in targeted attacks). We perform all 4 scenarios for targeted attacks, and settings \one~and \two~for untargeted attacks. 
In each scenario, we randomly select 50 videos from UCF-101 and HMDB-51 (25 from each), and use C3D as the target model. 

Table~\ref{tab_ablation_result_style} shows the StyleFool experimental results of the ablation study. 
Figure~\ref{Fig:visualization_ablation_targeted} in Appendix~\ref{appendix_exp_results} visualizes two video samples generated in targeted and untargeted scenarios. Note that the performance of StyleFool is reflected by not only the attack success rate and query number (Table~\ref{tab_ablation_result_style}), but also the naturalness and realness (Figure~\ref{Fig:visualization_ablation_targeted}). Only when all the metrics are satisfied, we consider an attack effective. 
Recall that StyleFool can reduce a large number of queries and thus ensure the attack success rate up to 100\%. All the attack success rates in the ablation study are 100\% since all attacks succeed within the query limit.
In targeted scenarios, if the style is selected randomly, not only the adversarial video is easy to be distinguished (\eg weird colors and textures in Figure~\ref{Fig:visualization_ablation_targeted}(b)), but also the query number is not reduced. 
Although color theme proximity brings higher similarity and indistinguishability in the stylized video, more queries are needed if we only consider the color theme proximity, since the output distribution of the classifier does not experience a major change after transfer. 
When only considering target class confidence, the number of queries can be reduced as stylized videos can be moved closer to the decision boundary. The indistinguishability, however, is not fairly satisfying (\eg the blue leaves in Figure~\ref{Fig:visualization_ablation_targeted}(d)). 
Significantly, when the two criteria are both considered in style selection, the adversarial videos have high sensory comfort and the attack is efficient. 
In untargeted scenarios, similar to targeted attacks, stylized videos have higher sensory comfort when color theme proximity is considered 
(observed from Figures~\ref{Fig:visualization_ablation_targeted}(g) and (h)).
Note that the average queries are also guaranteed to be about 25\% less than random selection. 

\begin{figure*}
    \centering
    \includegraphics[width=0.75\linewidth]{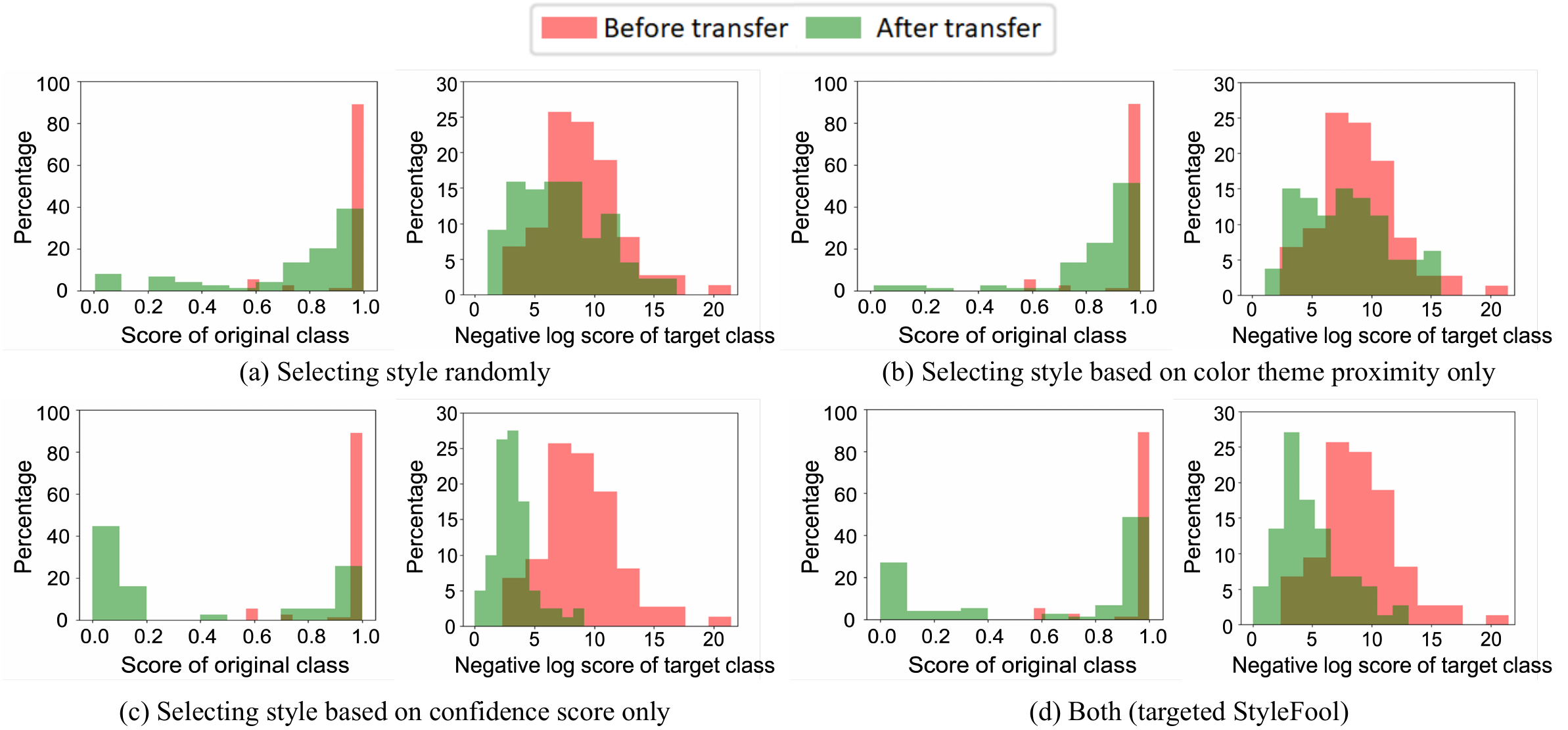}
    \caption{Confidence scores of the original class and target class before and after style transfer with different style selection strategies.}
    \label{fig:ablation_style_transfer}
\end{figure*}

\noindent \textbf{Contribution of style transfer in StyleFool.} 
In order to quantitatively evaluate the contribution of style selection and style transfer in StyleFool, we further compare the confidence scores before and after the style transfer process, \ie the original video in original (target) class vs. the stylized video in original (target) class.

Figure~\ref{fig:ablation_style_transfer} shows the confidence scores of samples in four style selection scenarios mentioned in the ablation study.
According to the scores of original class on the left side, before transfer, most original samples have confidence scores of 1.0 in their original classes. The scores decrease significantly after style transfer, which means that style transfer can actively bring the video sample close to or even across the decision boundary. In order to show the change of target class scores after style transfer, we present the distribution in a negative log score manner. The decrease of negative log scores after the style transfer indicates that the stylized videos have been moved closer to the decision boundary of the target class. Even though the style transfer cannot make every stylized video classified as the target class (\ie make the target class as top-1 score), such a move toward decision boundary will lead to the reduction of queries in the subsequent adversarial generation and boost the efficiency of attack. 
It is worth noting that, in some cases, stylized video samples can mislead the classifiers and succeed in the untargeted attack even without the adversarial sample generation process, as the confidence score of original class has already been significantly reduced and is no longer predicted as the top-1 score. No further queries are needed in such cases, \eg over 50\% of the stylized videos can successfully fool the C3D classifier under untargeted attacks for UCF-101, which provides a reasonable explanation that the minQ of StyleFool is 1 in Table~\ref{tab:attack_performance}.

\section{Countermeasures}\label{sec:countermeasures}
In this section, we show the performance of StyleFool against state-of-the-art video defenses and further discuss other mitigation strategies.

\subsection{Performance Against Defenses}
\noindent \textbf{Defense algorithms.~}
In order to further reflect the performance of StyleFool, we select three state-of-the-art video adversarial defenses {and one detection method} to evaluate the adversarial samples. 
\begin{itemize}[leftmargin=*]
    \item \textbf{AdvIT}~\cite{xiao2019advit} is the first video defense method, which uses optical flow to generate pseudo frames and evaluate the classifier output consistency between the pseudo frame and the target frame. As it is an adversarial frame detection method that only detects whether a video has been attacked, we use Area Under Curve (AUC) to measure its defense performance.
    \item \textbf{ComDefend}~\cite{jia2019comdefend} is a spatial defense method for videos, which has been claimed to have better defense performance against dense attacks, compared with the temporal defense~\cite{jia2019identifying}. 
    ComDefend employs a CNN (ComCNN) for compressing videos and another CNN (RecCNN) for video reconstruction.
    We use Defense Success Rate (DSR), the ratio of adversarial videos which are successfully defended, to evaluate the defense performance of ComDefend. Concretely, if the adversarial noise is successfully removed and the denoised video is correctly classified into the original class, the defense is considered as successful. 
    \item \textbf{RS}~\cite{jeremy2019certified} is a certified defense method in images, which utilizes noise sampled from Gaussian distributions to smooth the predicted scores and provides certifiably robust regions in $\ell_p$ norms to data samples~\cite{lee2019tight,jeremy2019certified,yang2020randomized}.
    RS has been proved to certify $\ell_2$ perturbations, but is not effective for $\ell_p \left ( p > 2 \right )$, especially $\ell_\infty$ perturbations~\cite{kumar2020curse,blum2020random}. We extend RS to videos and test the defense performance of video attacks. We use Clean Accuracy (CA), Adversarial Accuracy (AA), and Average Certified Radius (ACR) for defense evaluation. CA and AA indicate the prediction success rate of the smoothed classifier in the clean samples and adversarial samples, respectively. ACR represents the average value of the maximum certified radius of the smoothed classifier.
    \item {\textbf{CNN-generated Image Detector}~\cite{wang2020cnn-generated} is a detector for GAN-generated images, which trains a binary classifier to distinguish whether the image is generated or real. We extend CNN-generated image detection to videos to evaluate the performance of StyleFool, since StyleFool can be considered as a generative approach to some extent.}
    
\end{itemize} 

\begin{table}[t]  
\centering
\caption{AUC performance of AdvIT and DSR performance of ComDefend against adversarial attacks.}\label{tab_defense_advit_and_comdefend}
\resizebox{0.8\linewidth}{!}{
\begin{tabular}{cccccc}
\toprule
\multirow{2}{*}[-1ex]{\textbf{Model}} & \multirow{2}{*}[-1ex]{\textbf{Attack}} & \multicolumn{2}{c}{\textbf{AUC of AdvIT}} & \multicolumn{2}{c}{\textbf{DSR of ComDefend}} \\
\cmidrule(r){3-4} \cmidrule(r){5-6}
& & \textbf{UCF-101} & \textbf{HMDB-51} & \textbf{UCF-101} & \textbf{HMDB-51} \\
\midrule
\multirow{4}{*}{C3D} 
& Ilyas~\cite{ilyas2018black} & 99\% & 97\% & 85\% & 88\% \\
& V-BAD~\cite{jiang2019black} & 97\% & 94\% & 76\% & 78\%\\
& H-Opt~\cite{wei2020heuristic} & 93\% & 92\% & 74\% & 75\%\\
& StyleFool & \textbf{58\%} & \textbf{54\%} & \textbf{36\%} & \textbf{34\%}\\
\midrule
\multirow{4}{*}{I3D} 
& Ilyas~\cite{ilyas2018black} & 98\% & 95\% & 95\% & 93\% \\
& V-BAD~\cite{jiang2019black} & 96\% & 93\% & 87\% & 78\%\\
& H-Opt~\cite{wei2020heuristic} & 95\% & 93\% & 82\% & 75\%\\
& StyleFool & \textbf{53\%} & \textbf{52\%} & \textbf{53\%} & \textbf{48\%} \\
\bottomrule
\end{tabular}
}
\end{table}

In AdvIT, 50 adversarial videos are randomly selected from UCF-101 and 50 from HMDB-51. For AdvIT, 5 frames are selected from each adversarial video as target frames, and the average KL divergence is calculated as the consistency using optical flow between the target frame and its previous three frames. Columns 3 and 4 in Table~\ref{tab_defense_advit_and_comdefend} show the AUC performance of AdvIT against the four attack frameworks. StyleFool outperforms the other three benchmark frameworks with the lowest defense AUC. Although AdvIT utilizes optical flow to evaluate the temporal consistency of the adversarial video, StyleFool compromises AdvIT via considering the temporal loss in style transfer, which can increase the temporal consistency of stylized videos. In addition, since fewer queries are needed in the adversarial sample generation stage, the temporal consistency will not lose too much. 
Thus, it is not surprising that StyleFool performs better against AdvIT.

Columns 5 and 6 in Table~\ref{tab_defense_advit_and_comdefend} report the defense performance of ComDefend. In all scenarios, the defense success rates of StyleFool are lower than those of three benchmark frameworks, which indicates that ComDefend cannot turn adversarial videos from StyleFool back to clean videos. Specifically, when the target model is C3D, the defense success rates of Ilyas, V-BAD and H-Opt under two datasets are all higher than 74\%, while the defense success rates against StyleFool are less than 36\% (nearly half of 74\%). Even when attacking I3D, the defense success rates of StyleFool (both around 50\%) are still close to blind guess. Since ComDefend adds a Gaussian noise to the output of ComCNN, the adversarial perturbations can be removed after reconstructing the video in RecCNN. However, such a defense mechanism can only work for restricted perturbations. Since the mean of Gaussian noise added after ComCNN is 0, ComDefend is hard to defend against unrestricted perturbations.

\begin{table}[t]  
\centering
\caption{Performance of Randomized Smoothing against adversarial attacks. }\label{tab_defense_RS}
\resizebox{0.8\linewidth}{!}{
\begin{tabular}{cccccccc}
\toprule
\multirow{2}{*}[-1ex]{\textbf{Model}} & \multirow{2}{*}[-1ex]{\textbf{Attack}} & \multicolumn{3}{c}{\textbf{UCF-101}} & \multicolumn{3}{c}{\textbf{HMDB-51}} \\
\cmidrule(r){3-5} \cmidrule(r){6-8}
& & \textbf{CA} & \textbf{AA} & \textbf{ACR} & \textbf{CA} & \textbf{AA} & \textbf{ACR} \\
\midrule
\multirow{4}{*}{C3D} 
& Ilyas~\cite{ilyas2018black} & \multirow{4}{*}{66\%} & 52\% & 0.29 & \multirow{4}{*}{63\%} & 50\% & 0.29 \\
& V-BAD~\cite{jiang2019black} &  & 51\% & 0.29 &  & 49\% & 0.27 \\
& H-Opt~\cite{wei2020heuristic} &  & 44\% & 0.22 &  & 34\% & 0.19 \\
& StyleFool &  & \textbf{13\%} & \textbf{0.06} &  & \textbf{7\%} & \textbf{0.02}\\
\midrule
\multirow{4}{*}{I3D} 
& Ilyas~\cite{ilyas2018black} & \multirow{4}{*}{54\%} & 46\% & 0.26 & \multirow{4}{*}{56\%} & 46\% & 0.27 \\ 
& V-BAD~\cite{jiang2019black} &  & 45\% & 0.27 &  & 45\% & 0.24 \\
& H-Opt~\cite{wei2020heuristic} &  & 38\% & 0.21 &  & 41\% & 0.22 \\
& StyleFool &  & \textbf{9\%} & \textbf{0.04} &  & \textbf{7\%} & \textbf{0.03} \\
\bottomrule
\end{tabular}
}
\end{table}

\begin{table}[t]  
\centering
\caption{Performance of CNN-generated image detection on StyleFool. }\label{tab_defense_CNN}
\resizebox{0.60\linewidth}{!}{
\begin{tabular}{cccccccc}
\toprule
\textbf{Model} & \textbf{Dataset} & \textbf{Real Acc} & \textbf{Fake Acc} \\
\midrule
\multirow{2}{*}{C3D} 
& UCF-101 & 99.74\% & 27.89\% \\
& HMDB-51 & 99.38\% & 27.66\% \\
\midrule
\multirow{2}{*}{I3D} 
& UCF-101 & 99.76\% & 29.11\% \\
& HMDB-51 & 99.50\% & 26.88\% \\
\bottomrule
\end{tabular}
}
\end{table}

In RS, we set the number of Monte Carlo samples used for estimation as 10,000, and the variance of Gaussian noise as 0.25. The other parameters remain at their default values~\cite{jeremy2019certified}. 
We argue that a larger number of Monte Carlo samples is more convincing in video scenarios, due to the high dimension of video samples.
Table~\ref{tab_defense_RS} reports the defense performance of RS. In all scenarios, the clean accuracy exceeds 54\% (slightly lower than 65\% when applied in image scenarios~\cite{jeremy2019certified}). 
Due to the consideration of unrestricted perturbations in StyleFool, the adversarial accuracy is less than 13\%, and the average certified radius is less than 0.06, indicating that StyleFool can break the robustness certificate in a breeze. 
To conclude, adversarial examples crafted by StyleFool can escape from the robust regions induced by RS since StyleFool can conceal perturbations large in $\ell_p$ norms in unsuspicious patterns. To defend against StyleFool, RS is required to obtain larger robust regions with higher security budgets on the protected model, and this leads to a substantial utility trade-off, or even results in loss of utility.

Table~\ref{tab_defense_CNN} shows the experimental results of applying CNN-generated image detection on StyleFool. The method achieves near 100\% detection accuracy on real video samples (Real Acc), but less than 30\% detection accuracy on adversarial samples (Fake Acc).
The reason is perhaps because StyleFool introduces unrestricted perturbations in adversarial examples. Such a detection method that focuses on finding GAN-generated images without perturbations is not suitable.

\subsection{Other Mitigation Strategies}

\noindent \textbf{Adversarial training.~}
As one of the classical defenses against adversarial attacks, Adversarial Training (AT) can improve the accuracy of the classifier by training a mixture of clean samples and adversarial samples~\cite{goodfellow2014explaining,ali2019adversarial}. 
Adversarial training could be effective when the perturbations are restricted, as the possible perturbations can be effectively computed, resulting in a trained robust classifier. However, as mentioned previously, the unrestricted perturbations introduced by StyleFool have distinct distributions with that of restricted perturbations. Moreover, it is hard for the defender to determine possible perturbations since the perturbations can be diversified by choosing different style images. 
Therefore, the capability of adversarial training against StyleFool could be very limited. 

\noindent \textbf{Object detection with humans in the loop.~} 
With respect to the object detection in a video, although the output score of an object can be reduced much after StyleFool, human eyes still provide a high confidence for that object in the adversarial video. For example, a car in the adversarial sample may not be detected by an object detector due to StyleFool, but human eyes will always recognize the car even if its style has been transferred. Therefore, the adversarial videos can be distinguished from clean videos by human sensory cognition when the labels are inconsistent with the objects in the video. However, such mitigation will cost much and is even impossible for large-scale detection. 

\section{Discussion}\label{sec:discussion}

\noindent \textbf{Temporal consistency.~}
We guarantee the temporal consistency of the stylized video in the style transfer stage. Although irregular perturbations are generated in the adversarial attack stage, the detection results under AdvIT show that adversarial perturbations have little impact on the temporal consistency of adversarial videos. We also evaluate the SSIM between stylized videos and adversarial videos, as shown in Table~\ref{tab_video_quality_performance}. We find that the SSIMs under targeted attacks are all greater than 0.76 and those under untargeted attacks are all greater than 0.80, which indicate that videos before and after attacks have high similarity. It is worth noting that if the temporal loss in style transfer is added to the loss function in the adversarial attack stage, the consistency of the adversarial video may be further improved, and AdvIT will find it more difficult to detect the adversarial videos. However, such a strategy may increase the number of queries, since it requires the perturbations between two adjacent frames to be related.

\noindent \textbf{Online universal attacks.~}
Although StyleFool focuses on one-on-one offline attacks, our work can further support universal online attacks through certain changes. Similar to~\cite{huang2017real,gao2018reconet}, we can train a style transfer model for each style. When the input is a batch of videos, we only need to select the style image that can ensure sensory comfort according to the color theme proximity, and then input the video into the model corresponding to the style image to obtain the stylized video. Such procedure is similar to the previous work~\cite{li2019stealthy,xie2022universal} where the universal perturbation is first trained offline and then superimposed on the online video. We carry out a pilot study on a small batch of videos, and find that around 75\% of stylized videos can be directly misclassified (query is not allowed since it is online untargeted attacks), which is very close to the attack success rates of C-DUP and U3D (about 80\%). We will further explore the capability of applying StyleFool in the online attack scenarios in the future.

\noindent \textbf{Style selection for long videos.~} 
The state-of-the-art video classifiers involved in our research allow only 16-frame video inputs. The changes in the video frames with respect to colors and actions are small, as the duration of a 16-frame video is usually less than 1 second (played at 24 frames per second), which benefits the StyleFool quite a lot. 
We provide an analysis of the frame variation in Appendix~\ref{appendix_ssim}. 
For longer videos in the real world, the selection of the style image could be more complex, as in a long video, the action of the characters and the style of the scene may change frequently, and the performance of attack could be negatively influenced. A potential attack strategy could be extracting multiple style images at equal intervals and adding them to the style set as style images.

\noindent \textbf{Style detection.~} 
Various AI technologies are used to classify styled images~\cite{bar2014classification,sun2017convolution,cetinic2018fine-tuing,elvaigh2021gcnboost}. However, these style detectors can only classify pre-defined artistic styles, such as realism, romanticism, and symbolism. Considering that the styles involved in our study are all from real life with numerous (or even countless) styles, it is impossible to learn the style of each image through training, \eg mapping a specific video content to a specific style.
On the other hand, if we input the original video (or a frame) and an adversarial video into a style detector, it cannot tell which style is the original style and which one is generated, although the videos might be classified as two different styles. Further, the adversarial perturbations will also make it difficult to distinguish whether a video is stylized, since a small perturbation can greatly change the detection results.

\section{Conclusion}\label{sec:conclusion}
This paper mainly studies the video black-box adversarial attack, and proposes StyleFool against video classification systems. StyleFool generates unrestricted perturbations without changing semantic information, jumping out of the shackles of traditional restricted attacks. StyleFool selects style images based on the color theme proximity and the target class confidence, and transfers the initial video with the best style. Finally, NES is used for gradient estimation to solve the black-box setting. Experimental results show that the stylized videos can change the video style without changing the semantic content of the video, as well as push videos closer to or even across the decision boundary. StyleFool performs the best in both targeted and untargeted attacks. 
We also show that the adversarial videos generated by StyleFool are indistinguishable to human eyes, and the perturbations are observable but imperceptible. Finally, StyleFool can also evade the existing video defenses since most defenses are tailored for restricted perturbations. 
As a byproduct, we illustrate that the human-centric metrics developed in this paper can not only make stylized videos look natural to human eyes, but also make target class scores increase sharply, reducing the query number in adversarial attacks dramatically. In future work, we hope to explore possible new defenses against StyleFool.

\section*{Acknowledgment}
The authors are grateful to the anonymous reviewers for
their feedback that helped improve the paper. This work was supported in part by the National Natural Science Foundation of China (61972219), the Research and Development Program of Shenzhen (JCYJ20190813174403598, SGDX20190918101201696, 20210324120012033), the Overseas Research Cooperation Fund of Tsinghua Shenzhen International Graduate School (HW2021013), Guangdong Provincial Key Laboratory of Cyber and Information Security Vulnerability Research (No.2020B1212060081). Xi Xiao is the corresponding author of this paper.

\bibliographystyle{IEEEtran}
\bibliography{References}

\section*{Appendix}
\setcounter{section}{0}
\renewcommand{\thesection}{\Alph{section}}

\section{StyleFool Configuration}\label{appendix_configuration}
Following Huang~\etal~\cite{huang2017real}, the output of layer \texttt{Relu4\_2} is used in content loss, and the outputs of layers \texttt{\{Relu1\_1, Relu2\_1, Relu3\_1, Relu4\_1, Relu5\_1\}} are used for style loss. The selected layers are all replaceable, depending on personal preference.
We use DeepFlow~\cite{weinzaepfel2013deepflow} in optical flow estimation.

In style selection, we set bottom circle radius $r = 50$, height $h = 50\sqrt 3 $, weight coefficient $\mu  = 10^{4}$. 
In style transfer, we set weight coefficients $\alpha  = 10$, $\beta  = 75$ for targeted attacks and $\beta  = 50$ for untargeted attacks, $\gamma  = 10^{ - 3}$, $\lambda  = 10^{3}$. The weight coefficients $\mu $, $\alpha  $, $\beta $, $\gamma $, $\lambda $ are fine-tuned on 25 videos that are randomly selected from the UCF-101 dataset under C3D model. A grid search is carried on to find the most reasonable values. The current parameter combination is a reference with excellent performance. Users can slightly modify the parameters as needed. 
In gradient estimation, we set the number of Gaussian noise $n_g = 64$, the noise variance $\sigma = {10^{ - 6}}$ for targeted attacks, ${10^{ - 3}}$ for untargeted attacks. 

\section{Details of User Study}\label{appendix_user_study}
\noindent \textbf{Survey protocol.~}
Considering that the naturalness test and the realness test evaluate the indistinguishability of videos from different angles, it could be difficult for participants to make a fair enough decision in a short time period. In order to avoid bias, we evenly separate participants into 2 groups (80 participants each for the naturalness test and the realness test).
In the naturalness test, 20 randomly selected videos, including 10 adversarial samples and 10 clean videos, are presented to each participant. 
The participants are required to evaluate the naturalness after watching each video, and then judge the sensory comfort according to their cognition and common sense, based on a Likert scale~\cite{likert1932technique} from 1 to 5, representing ``very unnatural'', ``somehow unnatural'', ``neutral'', ``somehow natural'', and ``very natural'', respectively. 
In the realness test, 10 pairs of adversarial videos and their corresponding clean videos are displayed simultaneously to each participant who is asked to identify the realness of two videos (\ie which video is a clean video). Note that, participants are not informed of how many videos in each video pair are real, \ie we designed a multiple choice question with 5 answers, ``Video 1 real and Video 2 unreal'', ``Video 1 unreal and Video 2 real'', ``both real'', ``both unreal'', and ``unable to judge''. To filter out low quality or random responses, the answers that are made in less than 10 seconds (inclusive of the video playing time) are ignored. The average video duration is 7 seconds, and we have reserved at least 3 seconds for thinking.

\noindent \textbf{Detailed procedure.~}
We conducted the user study on the naturalness and realness aspects of the adversarial videos of StyleFool with
160 native speaker participants in total recruited on Amazon Mechanical Turk, a crowdsouring platform to hire remotely located ``crowdworkers'' to perform discrete on-demand tasks that computers are currently unable to do. It is operated under Amazon Web Services, and is owned by Amazon.
In our user study, we only hired participants over 18 years old who speak English as the first language and are from USA, UK and Australia with an answer approval rate of at least 95\%. A computer related background is not necessary for participants. We paid each participant an average of \$20.00/hr (\ie \$0.9 per question), which is higher than the average payment (\$11.00/hr) on the platform~\cite{hara2018data}.

\section{Additional Experimental Results}\label{appendix_exp_results}

\begin{figure*}[t]
    \centering
    \includegraphics[width=0.78\linewidth]{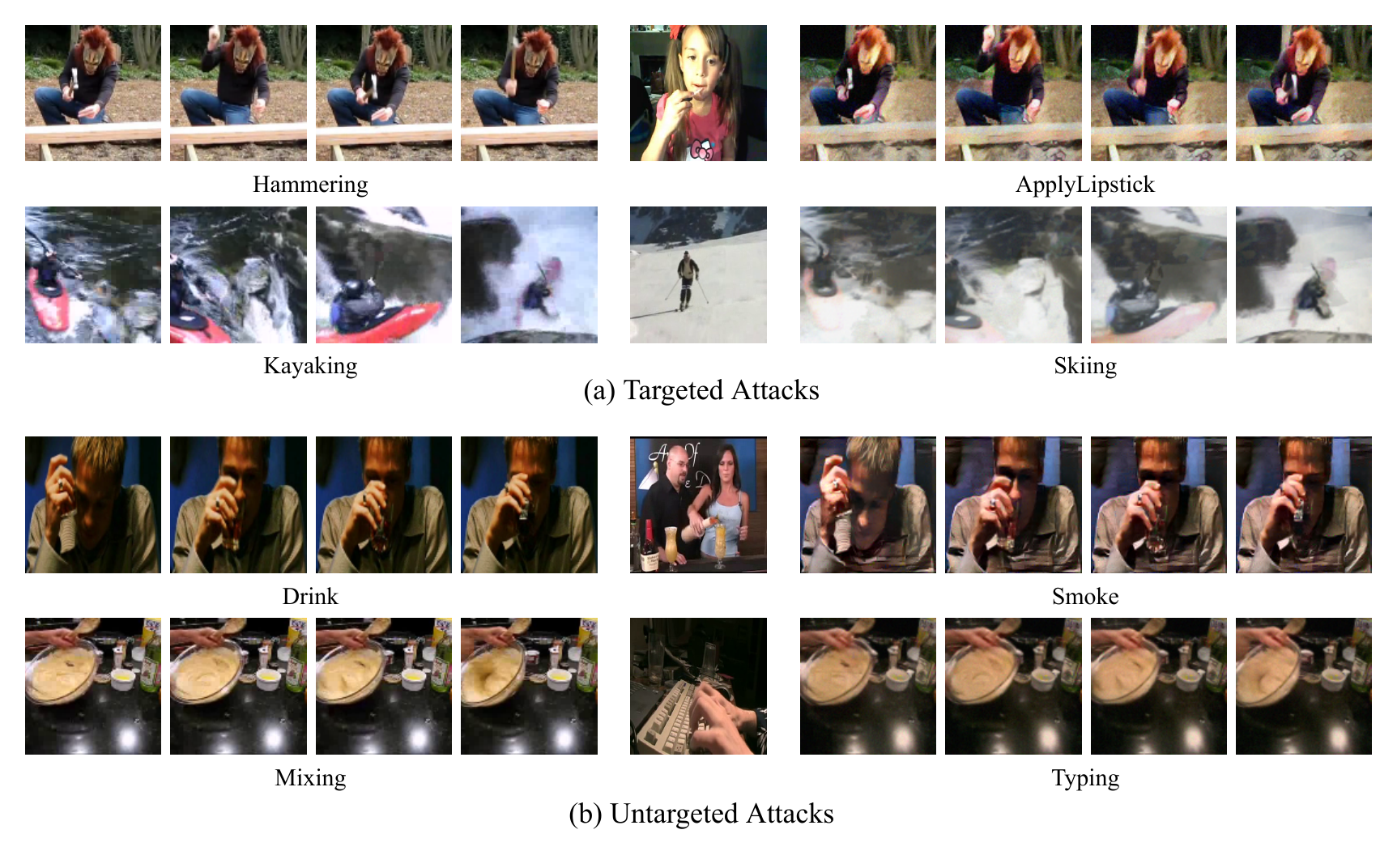}
    \caption{Visualization of StyleFool. The left four screenshots are taken from original videos, the right four screenshots are taken from adversarial videos, the image in the middle is the selected style image. }
    \label{Fig:visualization_StyleFool_simplify}
\end{figure*}

Figure~\ref{Fig:visualization_StyleFool_simplify} shows the original video samples and the corresponding adversarial video samples generated by StyleFool. Examples include both targeted and untargeted attacks. The styles of video samples are changed, but the semantic information remains unchanged. Please see more details in Section \ref{sec:experimental_results}. 

\begin{figure}[t]
    \centering
    \includegraphics[width=\linewidth]{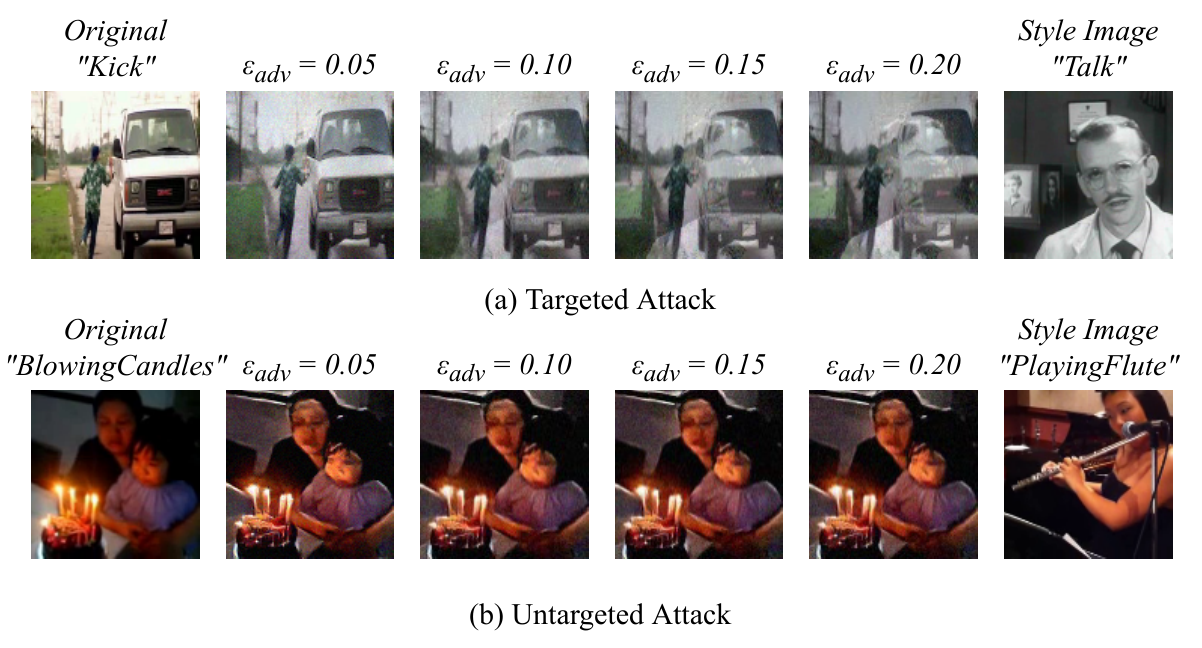}
    \caption{Visualization of StyleFool results with different $\varepsilon_{adv}$: (a) Targeted attack (``Kick'' to ``Talk''); (b) Untargeted attack (``BlowingCandles'' to ``PlayingFlute''). }
    \label{Fig:visualization_StyleFool_epsilon}
\end{figure}

\begin{figure*}[t]
\centering
\includegraphics[width=0.78\linewidth]{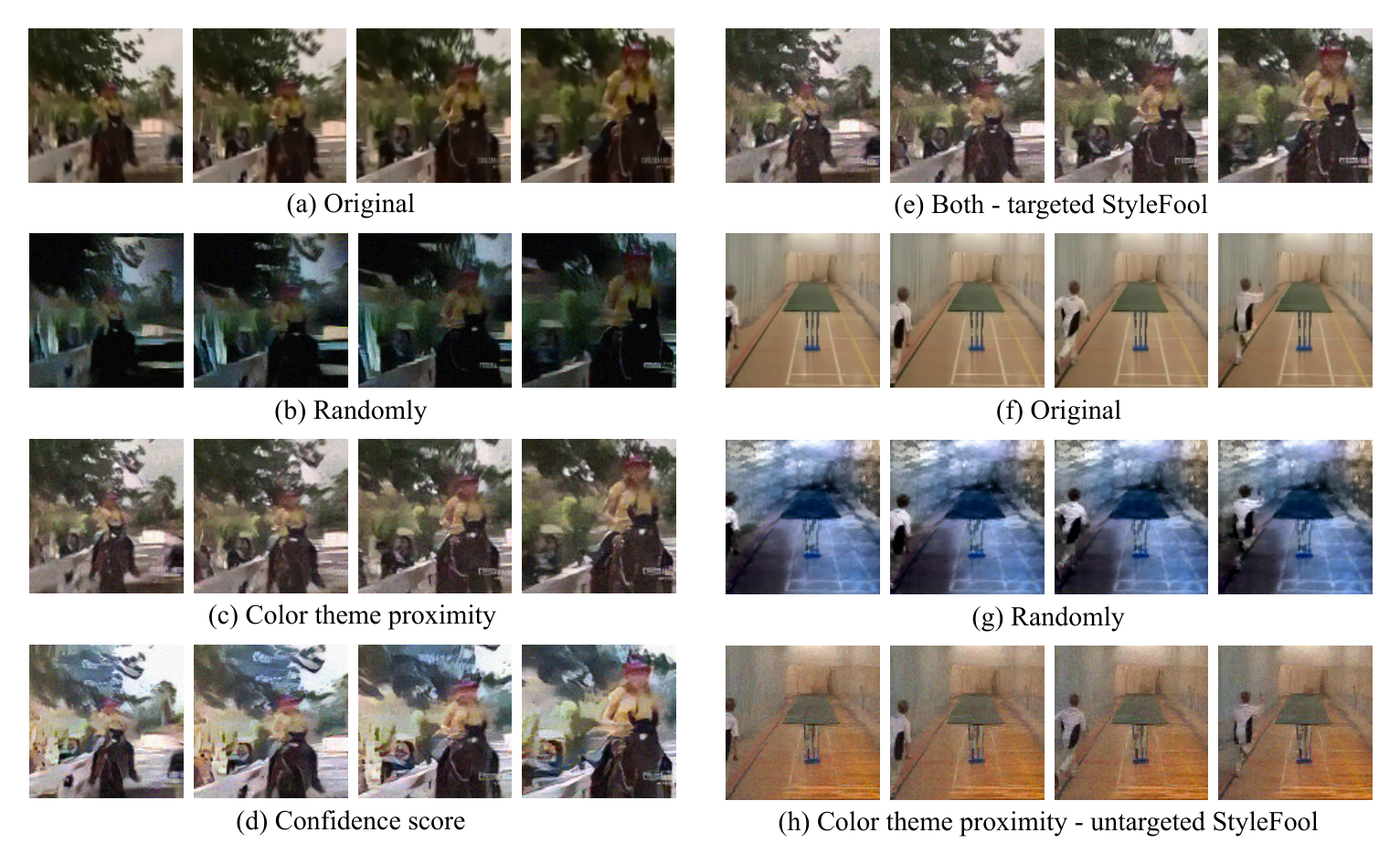}
\caption{Visualization of StyleFool results with different style selection strategies. (a)-(e) Targeted attacks; (f)-(h) Untargeted attacks. 
}\label{Fig:visualization_ablation_targeted}
\end{figure*}

\begin{figure}[t]
\centering
\includegraphics[width=\linewidth]{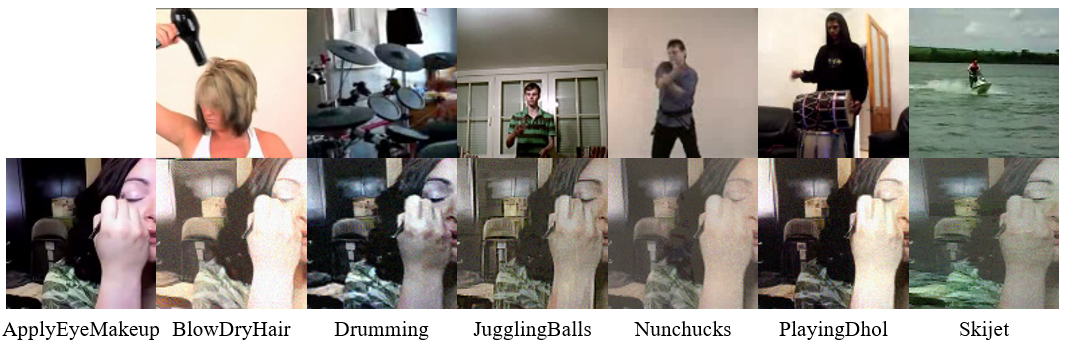}
\caption{Adversarial examples with different target labels for targeted attacks.
}\label{Fig:uncurated_targeted}
\end{figure}

We report the quantitative results of the indistinguishability in Table~\ref{tab_video_quality_performance}, including the SSIM, PSNR, and FID between original videos and adversarial videos (Ilyas~\cite{ilyas2018black}, V-BAD~\cite{jiang2019black}, H-Opt~\cite{wei2020heuristic} and ``StyleFool-ori-adv''), or between stylized videos and adversarial videos (``StyleFool-sty-adv''). Please see more details in Section \ref{sec:experimental_results}.

Figure~\ref{Fig:visualization_StyleFool_epsilon} shows the original video samples and the corresponding adversarial video samples generated by StyleFool under different $\varepsilon_{adv}$. Examples include both targeted and untargeted attacks. Please see more details in Section \ref{sec:ablation}.

Figure~\ref{Fig:visualization_ablation_targeted} shows the original video samples and the corresponding adversarial video samples generated by StyleFool under different style selection methods in the ablation study. Examples include both targeted and untargeted attacks. Please see more details in Section \ref{sec:ablation}.

\begin{figure}[t]
\centering
\includegraphics[width=\linewidth]{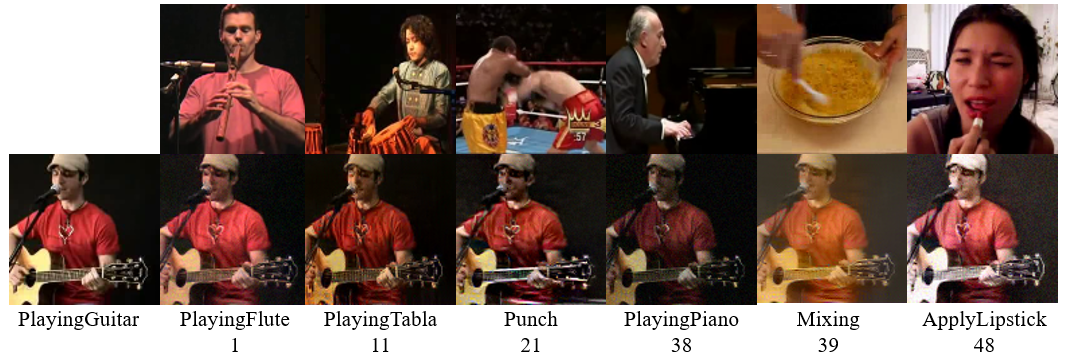}
\caption{Adversarial examples with different styles for untargeted attacks.
}\label{Fig:uncurated_untargeted}
\end{figure}

Figures~\ref{Fig:uncurated_targeted} and~\ref{Fig:uncurated_untargeted} provide more visualization results. In Figure~\ref{Fig:uncurated_targeted}, different target labels are selected in targeted attacks. The original video clip is shown in the first column. The style images, shown in the first row, are selected according to the criterion in Equation \ref{equ:targeted_criteria}. The corresponding adversarial video clips are shown below the style images.

In Figure \ref{Fig:uncurated_untargeted}, different styles are selected in untargeted attacks. Images in the first row represent the selected style images, while images in the second row stand for the original video clip and adversarial video clips. The number below the label name represents the ranking of the criterion when selecting the style. Please note that the ranking is not a continuous number because similar style images are hidden in order to visualize more adversarial examples. Note that, for untargeted attacks, only the sample with the smallest criterion has the best effect.

\begin{table*}[htp]  
\centering
\caption{Quantitative analysis on video indistinguishability.}
\label{tab_video_quality_performance}
\resizebox{0.9\linewidth}{!}{
\begin{tabular}{ccrrrrrrrrrrrr}
\toprule
\multirow{2}{*}[-0.5ex]{\textbf{Model}} & 
\multirow{2}{*}[-0.5ex]{\textbf{Attack}} & \multicolumn{3}{c}{\textbf{UCF-101 (Targeted)}} & \multicolumn{3}{c}{\textbf{HMDB-51 (Targeted)}}  & \multicolumn{3}{c}{\textbf{UCF-101 (Untargeted)}} & \multicolumn{3}{c}{\textbf{HMDB-51 (Untargeted)}}\\
\cmidrule(r){3-5}\cmidrule(r){6-8}\cmidrule(r){9-11}\cmidrule(r){12-14}
& & \footnotesize{\textbf{SSIM$\uparrow$}} & \footnotesize{\textbf{PSNR$\uparrow$}} & \footnotesize{\textbf{FID$\downarrow$}} & \footnotesize{\textbf{SSIM$\uparrow$}} & \footnotesize{\textbf{PSNR$\uparrow$}} & \footnotesize{\textbf{FID$\downarrow$}} & \footnotesize{\textbf{SSIM$\uparrow$}} & \footnotesize{\textbf{PSNR$\uparrow$}} & \footnotesize{\textbf{FID$\downarrow$}} & \footnotesize{\textbf{SSIM$\uparrow$}} & \footnotesize{\textbf{PSNR$\uparrow$}} & \footnotesize{\textbf{FID$\downarrow$}}  \\
\midrule
\multirow{5}{*}{C3D} & Ilyas~\cite{ilyas2018black} & 0.7610 & 27.6885 & 136.4496 & 0.7945 & 27.5548 & 122.5155 & 0.7895 & 30.7324 & 112.7122 & 0.7910 & 30.5597 & 107.1496 \\
& V-BAD~\cite{jiang2019black} & 0.7863 & 28.4952 & 119.1019 & 0.7948 & 29.3521 & 113.6902 & 0.7783 & 30.5321 & 108.0545 & 0.7873 & 30.5421 & 87.4092 \\
& H-Opt~\cite{wei2020heuristic} & 0.6768 & 23.4829 & 117.7430 & 0.7388 & 26.0329 & 109.1304 & 0.7144 & 30.6636 & 104.3397 & 0.7972 & 37.4704 & 102.6211 \\
& StyleFool-ori-adv & 0.4142 & 12.9681 & 215.6654 & 0.4149 & 12.8772 & 205.9897 & 0.5174 & 16.4794 & 156.4410 & 0.5786 & 18.8988 & 128.8778 \\
& StyleFool-sty-adv & \textbf{0.8653} & \textbf{29.4112} & \textbf{73.2175} & \textbf{0.8496} & \textbf{30.9609} & \textbf{97.9852} & \textbf{0.8735} & \textbf{45.2092} & \textbf{53.7769} & \textbf{0.9262} & \textbf{64.9507} & \textbf{30.7555}  \\
\midrule
\multirow{5}{*}{I3D} & Ilyas~\cite{ilyas2018black} & 0.6690 & 29.6478 & 136.0133 & 0.7100 & 29.5331 & 125.4818 & 0.6851 & 30.6850 & 124.6638 & 0.7075 & 30.5364 & 110.9063 \\
& V-BAD~\cite{jiang2019black} & 0.7199 & 29.2051 & 148.0453 & 0.7252 & 29.0609 & 136.1517 & 0.6861 & 30.5674 & 149.1361 & 0.6941 & 30.5106 & 132.6454 \\
& H-Opt~\cite{wei2020heuristic} & 0.7044 & 29.7635 & 163.5481 & 0.7255 & 28.5815 & 128.9625 & 0.7262 & 29.3567 & 141.4572 & 0.7742 & 31.6156 & 126.8154 \\
& StyleFool-ori-adv & 0.4009 & 13.4977 & 216.6966 & 0.4060 & 14.9125 & 222.1958 & 0.5113 & 16.3142 & 194.2943 & 0.4865 & 17.6216 & 187.8469 \\
& StyleFool-sty-adv & \textbf{0.7773} & \textbf{31.2119} & \textbf{104.2144} & \textbf{0.7607} & \textbf{30.3052} & \textbf{102.2426} & \textbf{0.8464} & \textbf{42.4192} & \textbf{102.1841} & \textbf{0.8092} & \textbf{46.6260} & \textbf{88.9165} \\
\bottomrule
\end{tabular}
}
\end{table*}

\begin{table}[htp]  
\centering
\caption{Preparation time cost per image. }\label{tab_time_preparation}
\resizebox{0.95\linewidth}{!}{
\begin{tabular}{cccccccc}
\toprule
\textbf{Generating styles} & \textbf{Calculating color proximity} & \textbf{Calculating confidence score} & \textbf{Total} \\
\midrule
0.0005s & 0.3864s & 0.0305s & 0.4174s \\
\bottomrule
\end{tabular}
}
\end{table}

\begin{table}[htp]  
\centering
\caption{Attack time cost. }\label{tab_time_attack}
\resizebox{1\linewidth}{!}{
\begin{tabular}{cccccccc}
\toprule
\multirow{2}{*}{\textbf{Model}} & \multirow{2}{*}{\textbf{Attack}} &  \multicolumn{4}{c}{\textbf{Time costs (min)}} \\
\cmidrule(r){3-6}
& & UCF101 (Targeted) & HMDB51 (Targeted) & UCF101 (Untargeted) & HMDB51 (Untargeted) \\
\midrule
\multirow{4}{*}{C3D} 
& Ilyas~\cite{ilyas2018black} & 105 & 74 & 17 & 7 \\
& V-BAD~\cite{jiang2019black} & 38 & 59 & 6 & 15 \\
& H-Opt~\cite{wei2020heuristic} & 188 & 224 & 20 & 53 \\
& StyleFool & 10 & 10 & 2 & $<$1 \\
\midrule
\multirow{4}{*}{I3D} 
& Ilyas~\cite{ilyas2018black} & 66 & 58 & 18 & 6 \\
& V-BAD~\cite{jiang2019black} & 32 & 28 & 16 & 5 \\
& H-Opt~\cite{wei2020heuristic} & 46 & 53 & 28 & 20 \\
& StyleFool & 10 & 11 & 1 & $<$1 \\
\bottomrule
\end{tabular}
}
\end{table}

Table~\ref{tab_time_preparation} shows the time cost in style set preparation. We also provide the attack time costs of StyleFool and benchmark attack methods with the same set of 100 randomly-selected videos in Table~\ref{tab_time_attack}. However, the number of queries is the main evaluation criterion for black-box adversarial attacks including StyleFool. The time cost is not a major concern as long as the model can be attacked through fewer queries. Please note that StyleFool requires extra time (about 21-28 mins) for style transfer.

Style transfer has demonstrated substantial efficacy in the search of the potential decision boundary of the target model. We compare the boundary search using style transfer and that with a surrogate model. Since we are the first one to propose unrestricted perturbations in videos, we can only compare style transfer with models with restricted perturbations. We randomly choose 100 videos from the dataset, and attack them using V-BAD against I3D model. These adversarial videos are again attacked using the same adversarial samples generation method of StyleFool against C3D model. As shown in Table~\ref{tab_surrogate_performance}, the results show that StyleFool can reduce at least half of the queries in both targeted and untargeted attacks, which means style transfer performs better in pushing videos to the decision boundary. The results also indicate that video adversarial samples lack transferability since large number of queries are still needed for attacks.

\begin{table}[t]
\centering
\caption{Attack performance of videos initiated by different methods.}
\label{tab_surrogate_performance}
\resizebox{0.95\linewidth}{!}{
\begin{tabular}{ccrrrrrrrr}
\toprule
\multirow{2}{*}[-0.5ex]{\textbf{Model}} & 
\multirow{2}{*}[-0.5ex]{\textbf{Attack}} & \multicolumn{4}{c}{\textbf{UCF-101 (Targeted)}} & \multicolumn{4}{c}{\textbf{HMDB-51 (Targeted)}}\\
\cmidrule(r){3-6}\cmidrule(r){7-10}
& & \footnotesize{\textbf{ASR}} & \footnotesize{\textbf{minQ}} & \footnotesize{\textbf{maxQ}} & \footnotesize{\textbf{AQ}} &  \footnotesize{\textbf{ASR}} & \footnotesize{\textbf{minQ}} & \footnotesize{\textbf{maxQ}} & \footnotesize{\textbf{AQ}} \\
\midrule
\multirow{2}{*}{C3D} & V-BAD+I3D & 100 & 2,538 & 288,454 & 130,116 & 100 & 205 & 272,999 & 75,756 \\
& StyleFool & \textbf{100} & \textbf{1,284} & \textbf{248,131} & \textbf{65,066} & \textbf{100} & \textbf{101} & \textbf{95,897} & \textbf{35,232} \\
\midrule
\multirow{2}{*}[-0.5ex]{\textbf{Model}} & 
\multirow{2}{*}[-0.5ex]{\textbf{Attack}} & \multicolumn{4}{c}{\textbf{UCF-101 (Untargeted)}} & \multicolumn{4}{c}{\textbf{HMDB-51 (Untargeted)}}\\
\cmidrule(r){3-6}\cmidrule(r){7-10}
& & \footnotesize{\textbf{ASR}} & \footnotesize{\textbf{minQ}} & \footnotesize{\textbf{maxQ}} & \footnotesize{\textbf{AQ}} &  \footnotesize{\textbf{ASR}} & \footnotesize{\textbf{minQ}} & \footnotesize{\textbf{maxQ}} & \footnotesize{\textbf{AQ}} \\
\midrule
\multirow{2}{*}{C3D} & V-BAD+I3D & 100 & 1 & 79,969 & 10,020 & 100 & 1 & 30,148 & 3,732 \\
& StyleFool & \textbf{100} & \textbf{1} & \textbf{19,013} & \textbf{3,811} & \textbf{100} & \textbf{1} & \textbf{9,948} & \textbf{1,526} \\
\bottomrule
\end{tabular}
}
\end{table}

\begin{table}[htp]  
\centering
\caption{SSIM analysis between frames. }\label{tab_SSIM_frames}
\resizebox{0.95\linewidth}{!}{
\begin{tabular}{cccccccc}
\toprule
\textbf{Frame index} & 1st and 2nd & 1st and the middle & 1st and the last \\
\midrule
\textbf{Average SSIM} & 0.5948 & 0.5411 & 0.5100 \\
\bottomrule
\end{tabular}
}
\end{table}

\section{HSV-to-XYZ Transformation}\label{appendix_hsv}
In the HSV space, \ie the HSV cone, let $r$ be the bottom radius and $h$ be the height of the cone. 
To compute the similarity between two pixels in the HSV space, we first convert an HSV coordinate $(H,S,V)$ to a coordinate $(X,Y,Z)$ in the XYZ space as follows:
\begin{equation}
\left\{ \begin{array}{l}
X = rVS\cos H\\
Y = rVS\sin H\\
Z = h\left( {1 - V} \right).
\end{array} \right.
\end{equation}
Given pixel values of the $i$-th color theme $(H^x_i,S^x_i,V^x_i)$ of the clean video $x$ and the $j$-th color theme $(H^s_j,S^s_j,V^s_j)$ of a style image $s$, we obtain their coordinates $\phi^x_{i}=(X^x_i,Y^x_i,Z^x_i)$ and $\phi^s_{j}=(X^s_j,Y^s_j,Z^s_j)$ in the XYZ space, respectively.

\section{SSIM Analysis between Different Frames}\label{appendix_ssim}
As discussed in Section \ref{sec:discussion}, we currently focus on 16-frame videos. Table~\ref{tab_SSIM_frames} shows an SSIM analysis between different frames, which indicates that the difference between the first frame and other frames in a single-action video is limited.

\end{document}